\documentclass[acmsmall]{acmart}
\usepackage{multirow}
\usepackage{cancel}
\usepackage{graphics}
\usepackage{makecell}
\usepackage{caption}
\usepackage{subcaption}
\usepackage{longtable}
\usepackage{amsmath}
\AtBeginDocument{%
 \providecommand\BibTeX{{%
 \normalfont B\kern-0.5em{\scshape i\kern-0.25em b}\kern-0.8em\TeX}}}

\setcopyright{acmcopyright}
\copyrightyear{2018}
\acmYear{2018}
\acmDOI{10.1145/1122445.1122456}

\acmJournal{JACM}
\acmVolume{37}
\acmNumber{4}
\acmArticle{111}
\acmMonth{8}

\begin{document}


\title{A Survey on Solving and Discovering Differential Equations Using Deep Neural Networks}
%

\author{Hyeonjung(Tari) Jung}
\email{jungx367@umn.edu}
\affiliation{%
\department{Electrical Engineering}
\institution{University of Minnesota}
\country{USA}
}
\author{Jayant Gupta}
\email{gupta423@umn.edu}
\affiliation{%
\department{Computer Science \& Engineering}
\institution{University of Minnesota}
\country{USA}
}
\author{Bharat Jayaprakash}
\email{jayap015@umn.edu}
\affiliation{%
\department{Mechanical Engineering}
\institution{University of Minnesota}
\country{USA}
}
\author{Matthew Eagon}
\email{eagon012@umn.edu}
\affiliation{%
\department{Mechanical Engineering}
\institution{University of Minnesota}
\country{USA}
}
\author{Harish Panneer Selvam}
\email{panne027@umn.edu}
\affiliation{%
\department{Mechanical Engineering}
\institution{University of Minnesota}
\country{USA}
}
\author{Carl Molnar}
\email{molna018@umn.edu}
\affiliation{%
\department{Computer Science \& Engineering}
\institution{University of Minnesota}
\country{USA}
}
\author{William Northrop}
\email{wnorthro@umn.edu}
\affiliation{%
\department{Mechanical Engineering}
\institution{University of Minnesota}
\country{USA}
}
\author{Shashi Shekhar}
\email{shekhar@umn.edu}
\affiliation{%
\department{Computer Science \& Engineering}
\institution{University of Minnesota}
\country{USA}
}


\renewcommand{\shortauthors}{}

\begin{abstract}
 Overview of Engineering 
\end{abstract}



\begin{abstract} 
Ordinary and partial differential equations (DE) are used extensively in scientific and mathematical domains to model physical systems. Current literature has focused primarily on deep neural network (DNN) based methods for solving a specific DE or a family of DEs. Research communities with a history of using DE models may view DNN-based differential equation solvers (DNN-DEs) as a faster and transferable alternative to current numerical methods. However, there is a lack of systematic surveys detailing the use of DNN-DE methods across physical application domains and a generalized taxonomy to guide future research. This paper surveys and classifies previous works and provides an educational tutorial for senior practitioners, professionals, and graduate students in engineering and computer science. First, we propose a taxonomy to navigate domains of DE systems studied under the umbrella of DNN-DE. Second, we examine the theory and performance of the Physics Informed Neural Network (PINN) to demonstrate how the influential DNN-DE architecture mathematically solves a system of equations. Third, to reinforce the key ideas of solving and discovery of DEs using DNN, we provide a tutorial using DeepXDE, a Python package for developing PINNs, to develop DNN-DEs for solving and discovering a classic DE, the linear transport equation.
\end{abstract}

\maketitle
\section{Introduction}
\label{sec:intro}
Forecasting the behavior of physical systems using a differential system often involves multiple challenges. First, the problem needs to be posed as a differentiable system, which for many domains is non-trivial. Complex systems often exhibit characteristics which are difficult or which make it difficult to solve any instance of a suitable model. Nonlinearity, stochasticity, and feedback loops are some example attributes of complex systems which make deriving and solving a mathematical model of physical observations challenging. The specific challenges encountered when attempting to solve each system are as diverse as the myriad number of mathematical models. This paper reviews a broad set of methodologies which focus on the applications of deep neural networks in the field of differential equations (DEs). 

The methodologies can be broadly categorized based on whether they solve forward problems or inverse problems. In solving a forward problem for a given system of differential equations (e.g, partial differential equation (PDE), ordinary differential equation (ODE)), the goal is to train DNN models which can mimic the solution to the input equations. For example, if a boundary value problem is known to accurately describe some physical system, the forward problem involves (approximately) solving that boundary value problem to determine the expected system behavior. An inverse problem involves, for a given dataset measuring key attributes of a physical system and identifying a system of differential equations (DEs) which can accurately explain the data. Solving for optimal coefficients for a given set of equations to describe observed data is perhaps the simplest example of an inverse problem.


Traditionally, the study of differential equations is performed by solving the equations through mathematical analysis (e.g. via reference point based optimization, element based optimization, etc.), often with the goal of approximating laboratory-based (physical) observations. In the last few decades, the study of differential equations using deep learning based (e.g., fluid dynamic simulations \cite{Raissi2017}, molecular modeling \cite{baker2019workshop}) software tools has gained traction as a sub-field of machine learning under the name of scientific machine learning (SciML) \cite{baker2019workshop}. Fueled by an increase in low-cost computational resources and rapid development of high level software libraries (e.g., Tensorflow, DeepXDE), the field is poised to grow rapidly. In this work, we focus on reviewing the use of deep neural networks for both discovering and solving differential equations.

DNN-based methodologies are important because they allow the use of vast computational resources (e.g., university computer clusters, commercial pay-per-use clusters) to improve understanding of differential equations. Historically, differential equations have laid the foundation of many disciplines and have been extensively used by engineering disciplines to model specific processes and phenomena (Table \ref{tab:PDEs}). 

\begin{table}[h]
\caption{Some important differential equations and their primary application domains. The classification follows the format: Type (ODE or PDE), Order, Linearity (Linear or Non-Linear).}
\centering
\resizebox{\columnwidth}{!}{%
\begin{tabular}{|l|l|l|l|l|}
\hline
Name &
  Developer (Year) &
  Classification &
  Application Domains &
  Equation \\ \hline
Bateman-Burgers equation &
  \begin{tabular}[c]{@{}l@{}}H. Bateman (1915)\\ M. Burgers (1948)\end{tabular} &
  PDE, 2, NL &
  \begin{tabular}[c]{@{}l@{}}Fluid mechanics\\ Transportation\end{tabular} &
  \begin{math}
      \begin{aligned}
        &\frac{\partial u}{\partial t}+u\frac{\partial u}{\partial x}=v\frac{\partial^2u}{\partial x^2}
      \end{aligned}
  \end{math}
  \\ \hline
Boltzmann equation &
  L. Boltzmann (1872) &
  PDE, 1, L &
  Thermodynamics &
  \begin{math}
      \begin{aligned}
        &\frac{\partial f}{\partial t}+\frac{\mathbf{p}}{m}\cdot\frac{\partial f}{\partial x}+{\mathbf{F}}\cdot\frac{\partial f}{\partial\mathbf{p}}=\left(\frac{\partial f}{\partial t}\right)_{coll}
      \end{aligned}
  \end{math}
  \\ \hline
Diffusion equation &
  A. Fick (1855) &
  PDE, 2, NL &
  \begin{tabular}[c]{@{}l@{}}Materials science\\ Thermodynamics\\ Heat transfer\\ Markov process\end{tabular} &
  \begin{math}
      \begin{aligned}
        &\frac{\partial\phi}{\partial t}=D\frac{\partial^2\phi}{\partial x^2}
      \end{aligned}
  \end{math}
  \\ \hline
Euler rotation equations &
  L. Euler (1765) &
  ODEs, 1, NL &
  Classical mechanics &
  \begin{math}
      \begin{aligned}
        &I\dot{\omega}+\omega\times\left(I\omega\right)=M
      \end{aligned}
  \end{math}
   \\ \hline
\begin{tabular}[c]{@{}l@{}}Euler-Bernoulli beam\\ equation\end{tabular} &
  \begin{tabular}[c]{@{}l@{}}L. Euler\\ J. Bernoulli (1750)\end{tabular} &
  ODE, 4, L &
  Solid mechanics &
  \begin{math}
      \begin{aligned}
        &\frac{\partial^2}{\partial x^2}\left(EI\frac{\partial^2w}{\partial x^2}\right)=q
      \end{aligned}
  \end{math}
  \\ \hline
Euler-Lagrange equations &
  \begin{tabular}[c]{@{}l@{}}L. Euler\\ J.-L. Lagrange (1755)\end{tabular} &
  PDE, 2, L &
  Classical mechanics &
  \begin{math}
      \begin{aligned}
        &\frac{\partial}{\partial t}\left(\frac{\partial\mathcal{L}}{\partial\dot{q}}\right)-\frac{\partial\mathcal{L}}{\partial q}=0
      \end{aligned}
  \end{math}
  \\ \hline
(Euler-)Tricomi equation &
  \begin{tabular}[c]{@{}l@{}}L. Euler\\ F. G. Tricomi (1923)\end{tabular} &
  PDE, 2, L &
  Fluid mechanics &
  \begin{math}
      \begin{aligned}
        &u_{xx}+xu_{yy}=0
      \end{aligned}
  \end{math}
  \\ \hline
Hamilton-Jacobi equation &
  \begin{tabular}[c]{@{}l@{}}W. Hamilton (1834)\\ C.Jacobi (1837)\end{tabular} &
  ODE, 1, NL &
  Classical mechanics &
  \begin{math}
      \begin{aligned}
        &H\left(\mathbf{q},\frac{\partial S}{\partial\mathbf{q}},t\right)+\frac{\partial S}{\partial t}=0
      \end{aligned}
  \end{math}
  \\ \hline
\begin{tabular}[c]{@{}l@{}}Hamilton-Jacobi-\\ Bellman equation\end{tabular} &
  \begin{tabular}[c]{@{}l@{}}W. Hamilton (1834)\\ C. Jacobi (1837)\\ R. Bellman (1957)\end{tabular} &
  PDE, 1, NL &
  \begin{tabular}[c]{@{}l@{}}Control systems\\ Dynamic programming\end{tabular} &
  \begin{math}
      \begin{aligned}
        &\frac{\partial V\left(x,t\right)}{\partial t}+\min_u{\left\{\frac{\partial V\left(x,t\right)}{\partial x}\cdot F\left(x,u\right)+C\left(x,u\right)\right\}}=0\\
        &V(x,t)=D(x)
      \end{aligned}
  \end{math}
  \\ \hline
Heat equation &
  J. Fourier (1822) &
  PDE, 2, L &
  Thermodynamics &
  \begin{math}
      \begin{aligned}
        &\frac{\partial u}{\partial t}=k\frac{\partial^2u}{\partial x^2}
      \end{aligned}
  \end{math}
  \\ \hline
Lotka-Volterra equations &
  \begin{tabular}[c]{@{}l@{}}A. J. Lotka (1920)\\ V. Volterra (1926)\end{tabular} &
  ODEs, 1, NL &
  Biological systems &
  \begin{tabular}[c]{@{}l@{}}
  \begin{math}
      \begin{aligned}
        &\frac{dx}{dt}=\alpha x-\beta xy\\
        &\frac{dy}{dt}=\delta xy-\gamma y
      \end{aligned}
  \end{math} \end{tabular} \\ \hline
Maxwell’s equations &
  J. C. Maxwell (1862) &
  PDEs, 1, NL &
  \begin{tabular}[c]{@{}l@{}}Electromagnetism\\ Optics\end{tabular} &
  \begin{tabular}[c]{@{}l@{}}
  \begin{math}
      \begin{aligned}
        &\nabla\cdot E=\frac{\rho}{\varepsilon_0\varepsilon_r}\\ 
        &\nabla\cdot B=0\\ 
        &\nabla\times E=-\frac{\partial B}{\partial t}\\ 
        &\nabla\times B=\mu_0\left(J+\varepsilon_0\frac{\partial E}{\partial t}\right)
      \end{aligned}
  \end{math}
  \end{tabular} \\ \hline
Navier-Stokes equations &
  \begin{tabular}[c]{@{}l@{}}C.-L. Navier (1822)\\ G. G. Stokes (1850)\end{tabular} &
  PDEs, 2, NL &
  Fluid mechanics &
  \begin{math}
    \begin{aligned}
      \rho\frac{Du}{Dt}=&\rho\left(\frac{\partial u}{\partial t}+u\cdot\nabla u\right)=-\nabla p+\\
                        &\nabla\cdot\left\{\mu\left(\nabla u+\left(\nabla u\right)^T-\frac{2}{3}\left(\nabla\cdot u\right)I\right)+\zeta\left(\nabla\cdot u\right)I\right\}+\rho g
    \end{aligned}
    
  \end{math}
  \\ \hline
Poisson’s equation &
  S.-D. Poisson (1813) &
  PDE, 2, L &
  \begin{tabular}[c]{@{}l@{}}Electrostatics\\ Fluid dynamics\end{tabular} &
  \begin{math}
    \begin{aligned}
      \nabla^2 f=g
    \end{aligned}
  \end{math}
  \\ \hline
Primitive equations &
  \begin{tabular}[c]{@{}l@{}}L. F. Richardson (1922)\\ W. M. Washington\\ C. L. Parkinson (1986)\end{tabular} &
  PDEs, 1, NL &
  \begin{tabular}[c]{@{}l@{}}Atmospheric science\\ Hydrodynamics\end{tabular} &
  \begin{tabular}[c]{@{}l@{}}
  \begin{math}
      \begin{aligned}
        &\frac{Du}{Dt}=\frac{\partial u}{\partial t}+u\frac{\partial u}{\partial x}+v\frac{\partial u}{\partial y}+w\frac{\partial u}{\partial p}=-\frac{\partial\phi}{\partial x}+fv\\ 
        &\frac{Dv}{Dt}=\frac{\partial v}{\partial t}+u\frac{\partial v}{\partial x}+v\frac{\partial v}{\partial y}+w\frac{\partial v}{\partial p}=-\frac{\partial\phi}{\partial y}+fu\\ 
        &0=-\frac{\partial\Phi}{\partial p}-\frac{RT}{p}\\ 
        &\frac{\partial u}{\partial x}+\frac{\partial v}{\partial y}+\frac{\partial w}{\partial p}=0\\ 
        &\frac{\partial T}{\partial t}+u\frac{\partial T}{\partial x}+v\frac{\partial T}{\partial y}+w\left(\frac{\partial T}{\partial p}-\frac{RT}{pc_p}\right)=\frac{J}{c_p}
      \end{aligned}
  \end{math}
  \end{tabular} \\ \hline
Road load equation &
  - &
  PDE, 2, NL &
  Transportation &
  \begin{math}
      \begin{aligned}
        P_{road}=\frac{1}{2}\rho C_DA\left(\frac{\partial x}{\partial t}\right)^3+C_{rr}Mg\frac{\partial x}{\partial t}+M\frac{\partial x}{\partial t}\frac{\partial^2x}{\partial t^2}+Mg\frac{\partial z}{\partial t}
      \end{aligned}
  \end{math}
  \\ \hline
Reynolds equation &
  O. Reynolds (1886) &
  PDE, 2, NL &
  Fluid mechanics &
  \begin{math}
      \begin{aligned}
        \frac{\partial}{\partial x}\left(\frac{\rho h^3}{12\mu}\frac{\partial p}{\partial x}\right)+\frac{\partial}{\partial y}\left(\frac{\rho h^3}{12\mu}\frac{\partial p}{\partial y}\right)=&\frac{\partial}{\partial x}\left(\frac{\rho h\left(u_a+u_b\right)}{2}\right)+\frac{\partial}{\partial y}\left(\frac{\rho h\left(v_a+v_b\right)}{2}\right)+\\
        & \rho\left(w_a-w_b\right)-\rho u_a\frac{\partial h}{\partial x}-\rho v_a\frac{\partial h}{\partial y}+h\frac{\partial p}{\partial t}
      \end{aligned}
  \end{math}
  \\ \hline
Schrodinger equation &
  E. Schrodinger (1926) &
  PDE, 2, L &
  Quantum mechanics &
  \begin{math}
      \begin{aligned}
        &i\hbar\frac{\partial}{\partial t}\Psi\left(x,t\right)=\left[-\frac{\hbar^2}{2m}\frac{\partial^2}{\partial x^2}+V\left(x,t\right)\right]\Psi(x,t)
      \end{aligned}
  \end{math} \\ \hline
Telegrapher’s equation &
  O. Heaviside (1876) &
  PDEs, 1, L &
  Electromagnetics &
  \begin{tabular}[c]{@{}l@{}}
  \begin{math}
      \begin{aligned}
        &\frac{\partial}{\partial x}V\left(x,t\right)=-L\frac{\partial}{\partial t}I\left(x,t\right)-RI\left(x,t\right)\\
        &\frac{\partial}{\partial x}I\left(x,t\right)=-C\frac{\partial}{\partial t}V\left(x,t\right)-GV\left(x,t\right)
      \end{aligned}
  \end{math} \end{tabular} \\ \hline
Wave equation &
  \begin{tabular}[c]{@{}l@{}}J.-B. d'Alembert (1747)\\ L. Euler (1760)\end{tabular} &
  PDE, 2, L &
  \begin{tabular}[c]{@{}l@{}}Electromagnetics\\ Acoustics\\ Fluid dynamics\end{tabular} &
  \begin{tabular}[c]{@{}l@{}}
  \begin{math}
      \begin{aligned}
        &1D: \frac{\partial^2u}{\partial t^2}=c^2\frac{\partial^2u}{\partial x^2}\\ 
        &3D: \nabla^2\psi=\frac{\partial^2\psi}{\partial x^2}+\frac{\partial^2\psi}{\partial y^2}+\frac{\partial^2\psi}{\partial z^2}=\frac{1}{c^2}\frac{\partial^2\psi}{\partial t^2}
      \end{aligned}
  \end{math} \end{tabular} \\ \hline
\end{tabular}%
}
\label{tab:PDEs}
\end{table}


This paper explains existing methods and summarizes the current literature on the use of DNNs to solve problems involving differential equations. We make no claim regarding the completeness of this review, though we believe it to be the most extensive and comprehensive review to date on the application of DNNs to solve DEs. The tutorial shows, with code snippets and output, how to use an existing open-source code package to solve two simple example problems -- one forward problem and one inverse problem -- involving the transport equation. Many of the findings from the literature review are discussed, along with future directions for related research. No strong conclusions are offered regarding the relative strength of numerical and DNN-based methods for solving specific DEs, though general advantages and drawbacks are identified for the emergent DNN-based methods based on problem characteristics. 

The paper is organized as follows. Section \ref{sec:motivation} explains why DEs are often difficult to solve, and how this motivates the use of DNN-DE solvers. 
Section \ref{sec:background} details the methods used to solve DEs, explaining the theory being utilized in many recent works, then it provides an organization of relevant problems involving DEs using a simple decision tree, and finally gives detailed descriptions of works related to problems in each node of the tree. Section \ref{sec:tutorial} provides a detailed computational example providing theoretical details for the forward problem using a basic network architecture and presents an instructive tutorial which provides a step-by-step example of how to solve a forward and an inverse problem, using open-source Python packages. Section \ref{sec:discussion} discusses several important topics that were mentioned in the literature or that follow directly from our review of the literature (e.g. regarding advantages and disadvantages of using DNNs to solve DEs). This section also further discusses the coding implementation used in the tutorial. Finally, Section \ref{sec:conclusions} concludes the paper outlines for related future work.







\section{Overview of Differential Equations and Mathematical Modeling} \label{sec:motivation} A partial differential equation models a physical process by defining relations between the partial derivatives of a multi-variable function. The partial derivatives describe changes in different variables which constitute the equation. DEs play an important role in engineering domains, where they are used to capture the dynamics of various systems. 
The concept of DEs as a mathematical model is a powerful tool to interpret physical observations, such as heat transfer or fluid movement. Once validated, the solution of DEs, in combination with methods for varying model parameters, can also be used to optimize the design of a device or process \cite{Comsol}.

\begin{figure}[htp]
 \centering
 \includegraphics[width=0.5\linewidth]{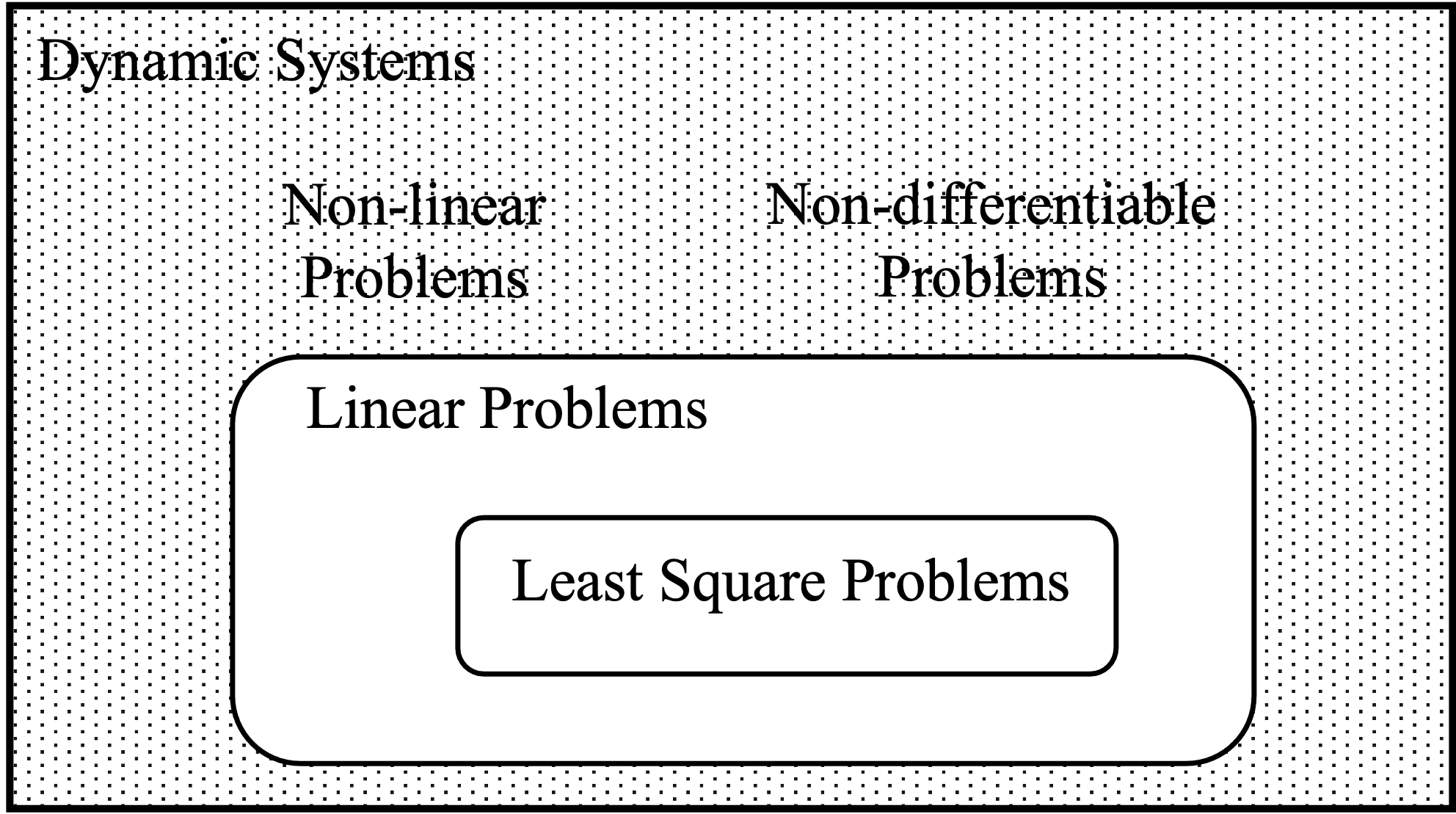}
 \caption{Categories of Mathematical Models; The paper focuses on Non linear problems and non-differentiable problems (as show in the Dotted background).}
 \label{Figure:mathsys}
\end{figure}

Traditional numerical methods for solving DEs (e.g. finite element methods) iterate through a finite number of states and find an optimized solution within the solution space. Depending on the complexity of the system and its solution space, these methods may not guarantee a single optimal solution. Finding and defining the constraints that yield a local, if not a global optimum is a unique process for each DE system. Figure \ref{Figure:mathsys} singles out linear systems within a broad categorization of dynamic systems. General method to solve linear problems, specifically least square problems, which are a subset among linear problems that can be solved using a single model, is described below. 



\subsubsection*{Least Square Problems:} A least square problem (LSP) is a problem modeled with no constraints and an objective which is a sum of squares of terms of the form $a_i^Tx-b_i$. 
\begin{equation} \label{eq1}
minimize\quad f_0(x) = ||Ax-b||^2_2 = \sum^k_{i=1}(a_i^Tx-b_i)^2 
\end{equation}
Problems that can be modeled in this form are generally solved effectively and reliably using existing methods. Extreme large problems, such as problems with millions of variables, or that require close to real-time solutions, will still run into challenges (e.g., scalability). 

\subsubsection*{Linear Problem:} Linear problems (LP) are problems where the objective and all constraint functions are linear:
\begin{equation} \label{eq2}
\begin{split}
 & minimize\quad c^Tx \\
 & subject\, to \quad a^T_ix \le b_i,\quad i=1,...,m
\end{split}
\end{equation}
Dantzig's simple method \cite{Nash2000} and interior point methods \cite{Nemirovski2004}, along with other existing methods, can reliably solve instances of LPs on a considerably large scale, although the challenge of exponential computational cost with added complexity still exists as in LSPs.



\subsubsection*{Other Dynamic Systems:} Systems outside of these classifications exhibit several characteristics that require examining whether a viable solution exists in any section of its solution space. Characteristics of such systems include:
\begin{itemize}
    \item The system is not convex: Convexity is a crucial characteristic that describes a system that is at least partially solvable. LPs and LSPs are both special cases of the general convex optimization problem \cite{Optimization1965}. Non-convex problems can be formulated as a convex problem with relaxed modeling (e.g., ensure that variables dependent on non-linear constraints are dropped, adjust constraints to only consider the subset of the problem that is convex). Modeling techniques ensure convexity is a rich area of study since each model influences appropriate techniques greatly.
    \item The system is non-linear: Non linear systems result in the formulation of an objective or constraint functions that are neither linear nor known to be convex. Similar challenges exhibited by non-convex systems hold true and no generalized solution can be inferred.
    \item The system includes high dimensions: If the domain of a system is in a high dimension, even if the problem is linear and has a global minimum, the computational cost to solve the system increases exponentially. This is one of the major challenges the conventional methods face when solving PDEs, since the multi-variable nature of PDE inherently implies the problem resides in a higher dimension.
\end{itemize}
\subsection{Why are DNN-DEs gaining attention?}
The historically influential equations listed in Table \ref{tab:PDEs} are classified based on the some of the characteristics discussed in Section \ref{sec:motivation}. Many of the systems of equations which are widely used in their respective fields are non-linear and involve high dimensional parameter spaces. Bypassing the arduous process of deriving a convex problem out of each system and efficiently solving models while avoiding an exorbitant computational cost remains as a central quest for virtually all engineering disciplines. A multitude of numerical methods have been and are being developed to solve DEs, but even state-of-the-art methods struggle to efficiently solve inverse problems and complex forward problems in high dimensions. Additionally, numerical methods must be applied to each instance of a problem separately, which can make prototyping and model tuning a burdensome process due to the computational expense of computing solutions for each model adjustment. With the rise of machine learning-based models, attempts to find generalized algorithms that can transfer learning of one system to the solving of another similar system is receiving optimistic attention. DNN-DEs, in particular, have shown promising results as a universal approximator that can act as both a linearized solver as well as an identifier of an unknown system. Recently it has been suggested that DNNs trained to solve DEs may counter common criticisms to quantify the generalization error \cite{Mishra2021}. Whether it can preserve the unique characteristics of a complex system while training on the defining patterns of its families of functions requires careful observation \cite{Berner2020,Hesthaven2018}, considering the diversity of systems from different domains each DNN-DE is trained on. Section \ref{sec:numerical_methods} further discusses the comparison between traditional numerical methods and DNN-DEs. The next section presents an overview of the literature on DNN-DEs and a taxonomy to help navigate the subject.

\section{Background, Classification of Methods, and a Taxonomy}
\label{sec:background}

The following sections are organized to provide a background on DNN-DE methods (Section \ref{sec:exploration}), a brief overview of physics-informed neural networks (Section \ref{sec:PINN_explain}), and a summary of literature focusing on these recent developments of DNN-DE methods. Key ideas of classifying the literature will be explained in Section \ref{sec:taxonomy} followed by detailed literature categorization in Section \ref{sec:taxonomy_org}.

\subsection{DNN-DE}
\label{sec:exploration}
A neural network (NN) can be considered as an optimization algorithm that iterates through a training data set and calculates the smallest gradient residual at each iteration. The network gained traction since multiple interconnected neural network layers can be stacked together to learn complex non-linear functions. Further, neural networks with a single hidden layer have been considered an equivalent to numeric solvers of PDE methods \cite{Hornik1989, Debao1993}. Furthermore, neural networks with a single hidden layer have been established as a universal approximator since the $1990$s. Deep neural networks attempt to leverage this approximation capability by using a significantly larger number of hidden layers, leading to the adjective `Deep'. The depth of these networks is an active field of research. Depending on the task and dataset, both the shallow (NN) networks (e.g., CIFAR 10 image recognition \cite{ba2014deep}) and deep (NN) networks (e.g., learning piece-wise smooth functions \cite{liang2016deep}) have been preferred.

Besides their depth, recent developments have demonstrated promising use cases of DNNs to solve DEs or metamodeling of DE-based systems. The advantages of using DNNs to approximate solutions of DEs can be summarized as follows: (i) DNNs are capable of expressing complex nonlinear relationships, mathematically supported by universal approximation theorems; (ii) forward evaluations of trained DNNs are extremely efficient, which is a desirable feature for real-time or many-query applications (e.g., in-flight ice detection \cite{zhou2019towards}); (iii) DNN models are analytically differentiable and thus derivative information can be easily extracted via automatic differentiation (Section \ref{App:AD}) for optimization and control problems.

The mechanism of training a DNN-DE can be either supervised or non-supervised. For supervised learning; the solutions of a known DE system become the training data. Raissi et al. \cite{Raissi2019} first demonstrated how to structure loss functions based on the governing DE residuals and training the DNN by minimizing the violation of DE constraints. Their paper examines several basic examples using well-known physics-based DE, such as Burgers' equation and Schr\"{o}dinger's equation. The method has been adopted in various engineering domains \cite{Han2020, Michoski2020, He2020}. 

\subsection{Physics-informed neural networks (PINNs) for solving DEs.} 
\label{sec:PINN_explain}
In this section, optimization of loss function using PINN is presented as a conceptual example of general DNN-DEs. The principle behind minimizing the loss function by manipulating the weights of the neural network can apply to various DNN-DE methods. 

Consider a DE parameterized by $\lambda$ for the solution $u(x)$ with $x = (x_1, \cdots, x_d)$ defined on a domain $\Omega \subset \mathbb{R}^d$:
$$f\left(x;\frac{\partial u}{\partial x_1}, \cdots, \frac{\partial u}{\partial x_d};\frac{\partial^2 u}{\partial x_1\partial x_1}, \cdots, \frac{\partial^2 u}{\partial x_1\partial x_d}; \lambda\right)=0, x \in \Omega,$$
with suitable boundary conditions
$$\mathcal{B}(u,x) = 0 \;\text{on}\; \partial \Omega, $$
where $$\mathcal{B}(u,x)$$ can be different boundary conditions (e.g., Dirichlet, Neumann, etc.).

We can describe PINN as a seuqence of 4 key steps \cite{Lu2019} as follows.
\begin{enumerate}
 \item Create a neural network $\hat{u}(x; \theta)$ with parameters $\theta$: In a PINN, a neural network $\hat{u}(x;\theta)$ is first constructed as a representative of the solution $u(x)$, which takes the input $x$ and outputs a vector with the same dimension as $u$. Here, $\theta =\{ W^{\ell} ,b^{\ell}\}_{1\leq \ell \leq L}$ is the set of all weight matrices and bias vectors in the neural network $\hat{u}$. One advantage of choosing a neural network as the surrogate of $u$ is that we can take the derivatives of $\hat{u}$ with respect to its input $x$ by applying the chain rule for differentiating the compositions of functions using automatic differentiation, a procedure which is conveniently integrated in machine learning packages such as TensorFlow\footnote{https://www.tensorflow.org/guide/autodiff} and PyTorch\footnote{https://pytorch.org/tutorials/beginner/blitz/autograd\_tutorial.html}.
 \item Develop two training sets for the equation ($\mathcal{T}_f$) and boundary/initial ($\mathcal{T}_b$) conditions: In this step, restrict the neural network $\hat{u}$ to satisfy the physics imposed by the PDE and boundary conditions. In practice, one can restrict $\hat{u}$ on some scattered points (e.g., randomly distributed points, or clustered points in the domain), i.e., the training data $\mathcal{T} =\{x_1, x_2,..., x_{|\mathcal{T}|} \}$ of size $|\mathcal{T}|$. In addition, $\mathcal{T}$ comprises of two sets, $\mathcal{T}_f \subset \Omega$ and $\mathcal{T}_b \subset \partial \Omega$ , which are the points in the domain and on the boundary, respectively. Refer to $\mathcal{T}_f$ and $\mathcal{T}_b$ as the sets of ``residual points". 
 
 \item Specify a loss function by summing the weighted $L^2$ norm of both the PDE equation and boundary condition residuals: To measure the discrepancy between the neural network $\hat{u}$ and the constraints, we define the loss function as the weighted sum of the $L^2$ norm of the residuals for the equation and the boundary conditions:
 $$\mathcal{L}(\theta; \mathcal{T}) = w_f\mathcal{L}_f(\theta;\mathcal{T}_f) + w_b\mathcal{L}_b(\theta;\mathcal{T}_b),$$
 where
 $$\mathcal{L}_f(\theta; \mathcal{T}_f) = \frac{1}{|\mathcal{T}_f|} \sum_{x\in\mathcal{T}_f} \left|\left| f\left(x;\frac{\partial u}{\partial x_1}, \cdots, \frac{\partial u}{\partial x_d};\frac{\partial^2 u}{\partial x_1\partial x_1}, \cdots, \frac{\partial^2 u}{\partial x_1\partial x_d}; \lambda\right)\right|\right|_2^2 ,$$
 $$\mathcal{L}_b(\theta; \mathcal{T}_b) = \frac{1}{|\mathcal{T}_b|} \sum_{x\in\mathcal{T}_b} ||\mathcal{B}(\hat{u},x)||_2^2$$
 and $w_f$ and $w_b$ are the weights. The loss involves derivatives, such as the partial derivative $\frac{\partial \hat{u}}{\partial x_1}$ for the normal derivative at the boundary $\frac{\partial \hat{u}}{\partial n} = \nabla \hat{u}\cdot n$, which are handled via automatic differentiation (Section \ref{App:AD}).
 \item In the last step, the procedure of searching for a good $\theta$ by minimizing the loss $\mathcal{L}(\theta, \mathcal{T})$ is called ``training''. Since the loss is highly nonlinear and non-convex with respect to $\theta$ \cite{blum1992training}, the loss function can be minimized by gradient-based optimizers such as gradient descent, Adam \cite{kingma2014adam}, and L-BFGS \cite{byrd1995limited}. Further, Lu et al. \cite{Lu2019} found that for smooth PDE solutions L-BFGS can find a good solution with fewer iterations than Adam, because L-BFGS uses second-order derivatives of the loss function, while Adam relies only on first-order derivatives. However, for stiff solutions L-BFGS is more likely to get stuck at a bad local minimum. The required number of iterations highly depends on the problem (e.g., the smoothness of the solution), and to check whether the network converges or not, we can monitor the loss function or the PDE residual using callback functions. They also noted that the training can be accelerated by using an adaptive activation function that may remove bad local minima; see \cite{jagtap2020adaptive}.
\end{enumerate}

\subsection{Classification Criteria}
\label{sec:taxonomy}
\begin{figure}[htp] 
 \centering
 \includegraphics[scale=0.6]{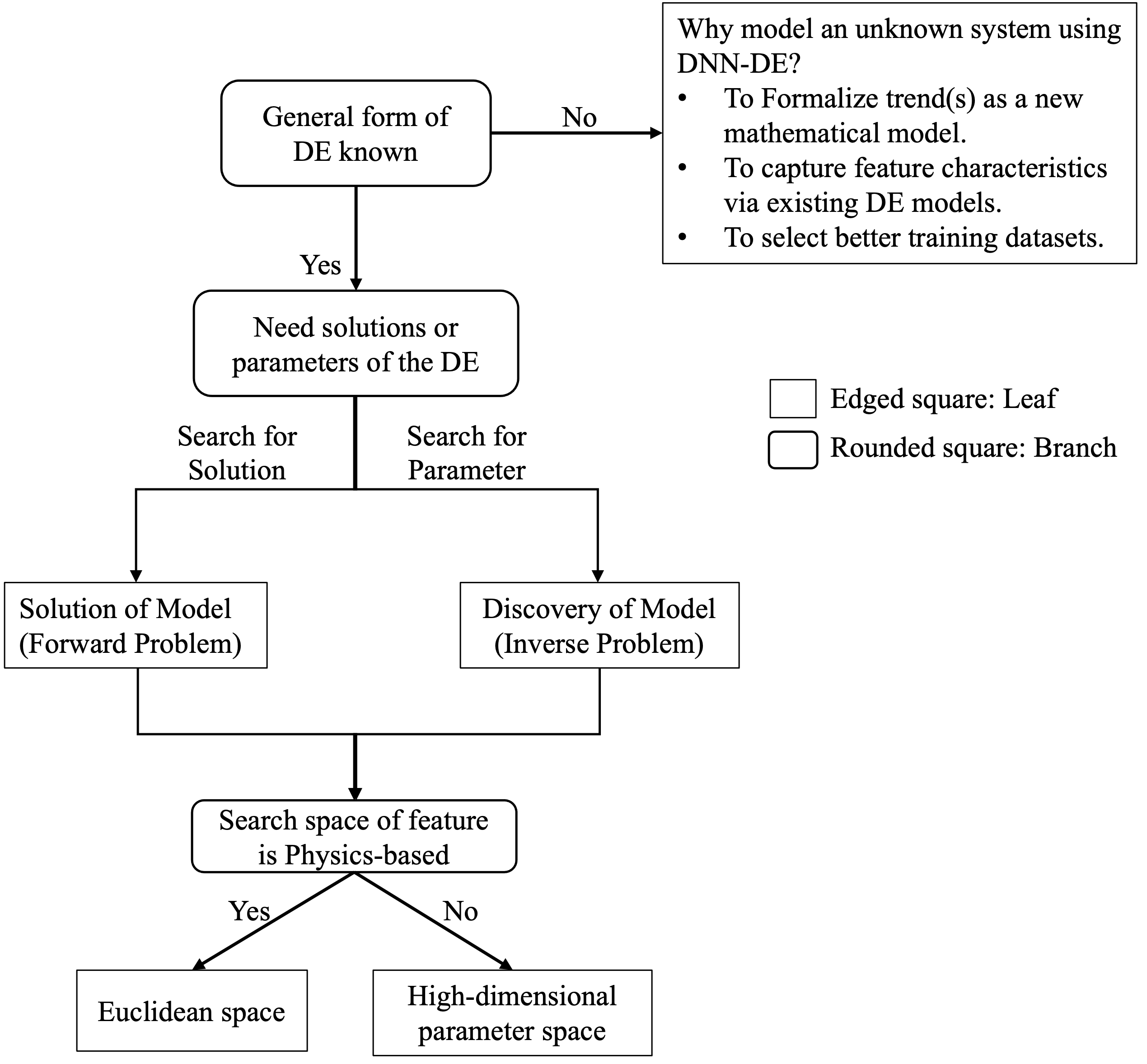}
 \caption{A taxonomy of DNN-DEs in the form of a decision tree}
 \label{fig:ont}
\end{figure}

In the following section, we present a taxonomy of recent works on designing or implementing DNN-DE methods. The taxonomy is organized as a decision tree for a reader whose application requires interpretation of a differential system. The first branch of the tree is split based on the assumption of working with a known family or known instances of differential equations. The branches define a classification which is further used to organize the literature in Section \ref{sec:taxonomy_org}. Organization of the taxonomy is presented as a decision tree in Figure \ref{fig:ont}. 

\subsubsection*{General formation of DE is available - Model Solving vs Model Discovery} The current trend of DNN-DE methods can be first categorized by two groups based on whether the objective of the trained neural network is to solve a single system of DEs or multiple DEs within the same family and to discover a single or multiple DEs. General formation of a system of DE becomes known when observations of a phenomenon are mathematically modeled. This mathematical model can be understood as a definite rule to describe the system through the progress of an independent variable, such as time or space. Therefore, for a known DE system, the main goal is to find a solution that satisfies all DEs of the system for a range of independent variables. When the DE system is unknown, discovering the system that best describes the observations becomes the challenge.

Most early work on DNN for DEs evaluate how well a DNN finds a solution for a fully discovered system of DEs. Traditionally neural networks were designed to learn mappings between finite-dimensional Euclidean spaces whereas, neural operators are designed to learn mapping between different functional spaces. The flexibility allows them to learn an entire DE family. The key idea behind the approach is to learn parameters which are mesh-agnostic i.e., the same set of parameters can be applied across multiple spaces. 

Raissi et al. \cite{Raissi2017} provide a foundational framework for posing discovery of a PDE via training DNNs. The paper demonstrates the derivation of continuous and discrete DE models as DNNs, along with mock-up examples of modeling linear equations. Because parameters of a DE map to boundary conditions of the DE, implementing methods to detect eligible parameters in the DNN-DE methods architecture becomes an important consideration. Berg and Nystr\"{o}m \cite{Berg2018} proposed a deep feed forward neural network implementation where the suggested optimal points from the network is explicitly designed to fulfill the constraint via distance calculation. 

Raissi et al.'s paper \cite{Raissi2019} was the first papers to propose physics-informed neural networks, which they describe as “neural networks that can solve supervised learning tasks while respecting any given laws of physics described by general nonlinear differential equations”. They aimed to set the foundation for a new approach to modeling and computation that utilizes deep learning in mathematical physics. They did this by dividing the paper into data-driven solutions of partial differential equations (also referred to as forward problems) and data-driven discovery of partial differential equations (also referred to as inverse problems). To demonstrate data-driven solutions of differential equations, they use the example of Schrodinger’s equation (continuous time model) and the Allen-Cahn equation (discrete-time model) to demonstrate data–driven discovery, they use the example of the Navier-Stokes equations (continuous-time model) and the Korteweg–de Vries equation (discrete-time model).

\subsubsection*{Search space of feature is Physics based - Euclidean space vs High-dimensional Parameter Space}
The functional space of a DE system defines the shape and space a function can take. Thus, a two-dimension function cannot capture a 3-dimensional phenomenon. Further, in this paper functional space also refers to the shape of functions that can be characterized together broadly. For instance, a polynomial function of $f(x) = x^2$ would have a different functional domain than $f(x)=x$. Often times a DE system will have a range where independent variables of the model stay valid. This solution space gets defined as a set of boundary conditions. 

\begin{figure}[htp]
 \centering
 \includegraphics[width=0.75\linewidth]{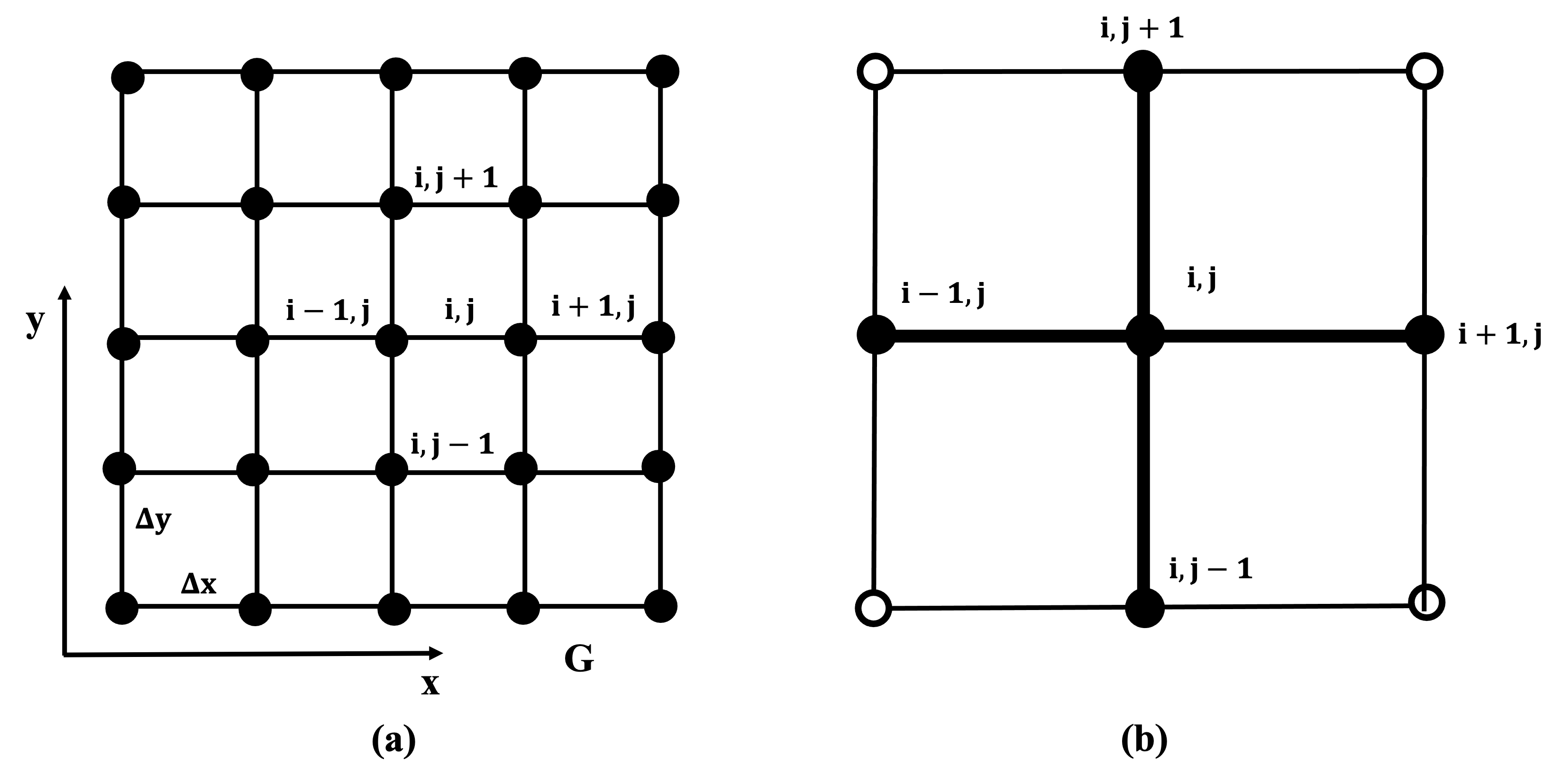}
 \caption{Illustration of a grid-like division of search space. (a) shows finite method's continuous definition of the grid where boundary and initial condition of each box is strictly imposed. (b) shows a single partition of the grid, which a DNN-DE with physics-based Model uses as a discrete label.}
 \label{Figure:grid}
\end{figure}

Finite numerical methods often define the solution space as a grid where each division can be solved as a convex function. Finding the appropriate size and shape of the grid becomes the key question when assessing the feasibility and flexibility of the optimization method. Although training a neural network does not require or guarantee convergence of any portion of its solution space, many DNN-DE methods have adopted the principle of segmenting the problem space. Figure \ref{Figure:grid} illustrates a visual example of how a space can be divided in a Euclidean grid, both in a finite volume method (a) and in DNN-DE method (b).  A number of research works reviewed in the following section have used the segmentation of functional space as to capture its learning features with finer granularity.

In this paper, we distinguish models in Euclidean space from models in high-dimensional parameter space. We classify papers with grid-like segmentation of its search space is what we will define as Euclidean Space learning. This distinction is clear in \cite{Raissi2018a} where a method to solve for the time-dependent DE model makes no segmentation of space thus, training the neural network agnostic of the search space outside each training data. However, the same architecture (i.e. presened in \cite{Raissi2018a}) was modified to consider the possibility of specifying residual points that effectively act as a grid-like structure where the trained DNN can more efficiently capture critical points of the trained functional space's characteristics \cite{Lu2021}. A later project that adopts the Physics-induced Neural Network architecture introduced in \cite{Raissi2018a}, makes a clear delineation of such strategy \cite{Portwood2019}. Leveraging convolution neural network (CNN) architecture, Gao et al. \cite{Gao2021} implemented geo-spatial aware DNN-DE that model physical systems mapped on actual locations. This is another example of physics based search space.

Examples of non-Physics space based models are provided here for a further clarification. Reinforcement learning used in \cite{e_deep_2017} where the value function of an agent's decision process is not formulated with assumptions regarding the functional families of the search space is another example of DNN-DEs with non-physics space based model. Vlachas et al. \cite{Vlachas2018} use long short-term memory (LSTM) architecture that first transforms a Physics-based dataset to a frequency domain via a Fourier transformation and uses its Fourier coefficients as the means to improve training accuracy, resulting the paper to be categorized as high-dimensional parameter space method. Li et al. \cite{Li2020c} applies a similar principle to Vlachas et al. \cite{Vlachas2018} but its neural network also considers the distribution of its Fourier coefficients in a spatial domain. 

\subsection{Taxonomy of Literature}
\label{sec:taxonomy_org}
In this section, literature exploring the field of DNN-DE are classified based on the criteria described in Section \ref{sec:taxonomy}.
\subsubsection*{Model Solving in Euclidean Space}
Samaniego et al. \cite{Samaniego2020} demonstrates solution driven DE methods implemented with a feed forward neural network. Their method was derived to solve boundary value problems of grid-like structured systems where the DNN-DE's solution would result in an approximated and optimized boundary of the given system. While most studies on parameter estimation use PDE systems of second-order, \cite{Samaniego2020} provides a Kirchhoff plate bending example of a fourth-order PDE. The work demonstrates that a higher order differentials can be posed as a transversal deflection of an equivalent second-order system. 

An example of a forward problem of engineering application, Portwood et al. \cite{Portwood2019} gives a practical account on modeling the flow dynamics of soil debris. A first order ODE system of turbulence was modeled with explicit physical constraints where its kinetic energy and dissipation rate parameters were trained using a recurrent neural network based architecture. The trained DNN-DE solutions were then validated against ground truth simulation data and numerical solutions. The results showed that DNN-DE based models were better for lower Reynold numbers, which correspond to the final stages of decay, than the analytical solver. This result was surprising since the parameterized Re number had generally been thought to correspond to higher variability in the model.

Another work on Deep Lagrangian Networks \cite{Lutter2019} explores training DNN-DE to solve for mechanic systems of Euler-Lagrange equation. The Lagrange equation, which describes the balance of kinetic and potential energy of a system as an ordinary differential equation is encoded in the DNN to find the network parameters that would minimize the imbalance. As the paper points out, constructing the imbalance between kinetic and potential energy as the independent variable to be optimized resembles the principle of PINN structure \cite{Raissi2018}.

There are various inter-disciplinary applications to physics informed neural networks (PINNs). For example, in \cite{misyris2020physics} PINNs are used to predict transient state behavior of a single bus infinite machine (SBIM) system. In particular, a system of equations (i.e., swing equation) is described to explain SBIM which are then learned by PINNs and can estimate system variables (e.g., rotor angle, frequency) and system parameters. In \cite{Stiasny2021}, a PINN based DNN-DE was modified with Runge-Kutta schemes to train on solving a similar SBIM DE set up. The empirical results showed significant improvement in the speed of the solution as well as comparable accuracy to the current numerical method, model reduction approach.


The Deep Ritz method is a deep learning based system \cite{E2017} can be used to solve high-dimensional variational problems that arise from partial differential equations. The examples used to demonstrate and validate the method are the Poisson equation, an equation with the Neumann boundary condition, and eigenvalue problems. The authors concluded that the treatment of essential boundary conditions is not as simple as in traditional methods.

\subsubsection*{Model Solving in High-dimensional Parameter Space} A paper by Vlachas \cite{Vlachas2018} explores the use of long-short-term-memory (LSTM) for data-driven forecasting of high dimensional systems. LSTMs are neural networks which are useful to learn long term dependencies in datasets \cite{hochreiter1997long}. In particular, they explore the use of LSTMs for learning turbulence equations such as the Kuramoto-Sivashinsky equation, and equations for different climate models. The results show that LSTM is able to learn the equations with different boundary conditions.

A ‘Deep Galerkin Method’ (DGM) algorithm, proposed in  \cite{Sirignano2018}, solves high dimensional PDEs by approximating the solution with a deep neural network while satisfying the initial conditions and the boundary conditions. The algorithm segments the training data into batches of randomly sampled time and space points and tests them on a variety of simple and complex differential equations including but not limited to a high-dimensional Hamilton-Jacobi-Bellman equation and Burgers’ equation. Various toy examples were tested in other sources to compare DGM's controllability and accuracy against equivalent or comparable analytical solutions or numerical approximations, such as Monte Carlo and Finite-differences \cite{AL-ARADI2018}.

In \cite{Berner2020}, a single deep neural network is used, trained on simulated data, to solve a family of high-dimensional linear Kolmogorov partial differential equations, which are mainly affected by the curse of dimensionality. The authors reformulate the numerical approximation of the family under consideration as a single statistical learning problem using the Feynman-Kac formula and validate the efficiency of the algorithm on the heat equation with varying diffusion coefficients and the Black-Scholes option pricing models. 

Another paper \cite{e_deep_2017} proposes a new algorithm for solving parabolic partial differential equations and backward stochastic differential equations (BSDEs) in high dimensions. The authors first formulated the PDEs as BSDEs using the nonlinear Feynman-Kac formula. These BSDEs were then solved as a model-based reinforcement learning problem with the gradient of the solution being the policy function. Finally, the policy function was then approximated by a deep neural network.

A new numerical method \cite{beck_deep_2021} approximates solutions of high dimensional nonlinear parabolic PDEs. The method splits the differential operator into a linear and a nonlinear part and uses deep learning together with the Feynman–Kac formula to iteratively solve linear approximations of the equation over small time intervals. This breaks the PDE approximation task into smaller problems that can be solved successively.

\subsubsection*{Model Discovery in Eucliedean Space}

Wu and Xiu \cite{Wu2020} presented a framework based on training deep neural networks for approximating unknown autonomous partial differential equations. The framework approximates the evolution operator of the underlying differential equation, which is an operator that maps the solution from a current timestamp to a future timestamp that completely characterizes how the solution of the underlying PDE evolves with time. The framework achieves a finite dimensional approximation by training a residual network based deep neural network. The author demonstrates the applicability of the framework on a set of examples including Burgers' equation, the diffusion equation, the advection equation and the 2-dimensional convection-diffusion equation.

A deep learning method by Raissi and Karniadakis \cite{Raissi2018} discovers nonlinear partial differential equations from scattered and noisy spatio-temporal datasets. They utilized 2 deep neural networks to approximate the unknown solution and the nonlinear dynamics of the underlying processes. Respectively, they validated the method on Burgers equation, the Korteweg-de Vries equation, the Kuramoto-Sivashinsky equation, the nonlinear Schrodinger equation and the Navier-Stokes equations. 

\cite{Gao2021} derives a parametric solution of a heat dissipation system using a CNN driven architecture. The work explores a multi-dimensional system of simulated nature where its numerical results were compared against the finite volume method. 

Mishra \cite{Mishra2018} implements a stochastic gradient descent method to target time-dependent ODEs and PDEs. The paper examines on a foundational level how concepts of linear and non-linear ODEs can be solved using simple linear oscillation and non-linear population growth systems as examples. One example of a PDE in the form of a heat equation was also presented. The work reported all accuracy and training results and noted the challenge of quantifying the degree of error between its prediction and the true solution of the DE system. 

Mishra and Molinaro extended the work in \cite{Mishra2018} to DNN-DE solvers based on \cite{Mishra2021} based in PINN architecture. The applications presented in this work all adopted Euclidean space based model and formalize the generalization error based on this assumption. Further discussion of this topic can be found in Section \ref{sec:generalization error}.

The authors in \cite{Hesthaven2018} approaches the generalization error of its DNN-DE by making close comparison to its training dataset, solutions of several version of the family of DEs produced via the finite element method (FEM). Formal comparisons each partition's deviation and aggregated relative errors between FEM, Deep Galerkin method, and PINN-like DNN method were presented. Recognizing and quantifying the translation error between solutions, the paper emphasizes DNN-DE's advantage of low computational cost while accounting for the potentially low fidelity of a DNN-DE discovered model.

\subsubsection*{Model Discovery in High-dimensional Parameter Space}

The problem of quantifying the generalization error addressed in \cite{Mishra2018} was further examined in \cite{Lye2020}. This work presents a likely workflow for discovering a complex engineering problem. Using a general fluid dynamics system as a main example, the paper first illustrates the exponential computational demand of forward solving such problem, such as iterative forward solving algorithms, which include finite difference, finite volume, and discontinuous Galerkin methods. Suggesting uncertainty quantification as an alternative, where a complex system is approximated as a constrained time-space grid, the paper points to deep neural network as a universal approximator to be used in the quantification. Quantifying and putting a bound on the generalization error, an error accounting for solving forward problems using an approximated function when the exact parameters of such function was not determined, is more thoroughly explored by the same research team in \cite{Mishra2021}. 

A paper by Yi \cite{yi_efficient_2020} aims to improve the efficiency of the normal equation method, which is an analytical approach to Linear Regression with a least square cost function and one of the most widely used deep learning algorithms. The author proposed an efficient architecture, which first involved a systolic gaussian elimination. Secondly, Yi proposed a systolic matrix inversion based on Gauss-Jordan elimination and based on these designs, Yi proposed an efficient method to solve normal equations in DL problems.

Li et al. \cite{li2020fourier} propose that convolution parameters be learned in Fourier, which are discretization-invariant. This allows to be learned mappings across arbitrary spaces. Neural operators have also been developed for graph neural networks (GNNs) \cite{li2020multipole}. The operators help to reduce the computational complexity in high complexity models, using techniques inspired by the fast multipole algorithm.

\subsubsection*{Modeling an unknown system using DNN-DE}
It is worth noting that there have been recent examples of modeling a system as it maps to an arbitrary system of differential equations. We list a few examples below.

Yan et al. recently \cite{yan_robustness_2021} provided the first systematic empirical study of the robustness of neural ODEs and conclude that they are more robust compared to conventional CNN models. They also proposed a time-invariant steady neural ODE (TisODE) method, which is simple yet effective in significantly boosting the robustness of neural ODEs. It does so by regularizing the flow on perturbed data via the time-invariant property and the imposition of a steady-state constraint. The authors showed that the proposed TisODE outperforms the vanilla neural ODE and also can work in conjunction with other state-of-the-art techniques to further improve the robustness of deep networks.

Wang \cite{wang_learning_2020} aimed to design new numerical schemes in an autonomous manner. The paper proposes viewing iterative numerical PDE solvers that solve 1-dimensional conservation laws as a Markov decision process and use in a reinforcement learning to find new and potentially better data-driven solvers. They introduced a numerical flux policy which was able to decide on how numerical flux should be designed locally based on the current state of the solution. 

Recent work by Xu \cite{xu_dl-pde_2021} is an example of using DNN-DEs to quickly build many training samples that are rich in features and grounded in physical laws. The authors propose a deep-learning based data-driven method, called DL-PDE, to discover the governing PDEs of underlying physical processes. In this method, for a physical problem, a deep neural network is trained using available data, large amounts of meta-data are generated, and the required derivatives are calculated by automatic differentiation. The form of the underlying PDE is then discovered using sparse regression. 

Lu et al's work \cite{lu_beyond_2018} aims to connect the architectures of popular deep networks and discretizations of ODEs which could enable the design of new and more effective deep networks. They introduced an effective linear multi-step architecture (LM-architecture) that can be used on any ResNet (Residual neural network) - like networks and is inspired by the linear multi-step method solving ordinary differential equations. They applied the LM architecture on ResNet and ResNeXt and achieve higher accuracy than the original networks on CIFAR (Canadian Institute for Advanced Research) and ImageNet (dataset of annotated photographs intended for computer vision research) with a comparable number of trainable parameters.

\section{A Detailed Computational Example}
\label{sec:tutorial}
To solidfy the key ideas of DNN-DE, we provide a basic example of solving PDEs using DeepXDE \cite{Lu2019}, a Python wrapper for physics informed neural network (PINN) architecture, for both a forward problem (solving for {u} given an equation) and an inverse problem (deducing an equation from the system's input). Mathematical exploration of Physics Informed Neural Network (PINN) method is demonstrated in Section \ref{App:PINN_EG} with node-wise calculation of a single layer neural network modeling a simple DE. In Section \ref{dnn_pde_example}, solution and discovery of model for a simple DE using the PINN architecture is demonstrated.

\subsection{Example of Physics Informed Neural Network to solve a PDE}
\label{App:PINN_EG}
We consider a PDE defined as $\frac{\partial u}{\partial t} + 3.\frac{\partial u}{\partial x} = 0$ with initial condition $u(x, 0) = x.e^{-1*x^2}$. The PINN models function $u(x,t)$ as $\hat{y}$, for which the loss function can be defined as a linear combination of the PDE and the initial condition as follows:

\textbf{Loss function} $\mathcal{L}$ = $\frac{1}{|T_f|}*\left|\left|\frac{\partial \hat{y}}{\partial t} + 3\frac{\partial \hat{y}}{\partial x}\right|\right|^2 + \frac{1}{|T_b|}||\hat{y}(x, 0) - x.e^{-1*x^2}||^2$

\begin{figure}[htp]
 \centering
 \includegraphics[width=0.75\linewidth]{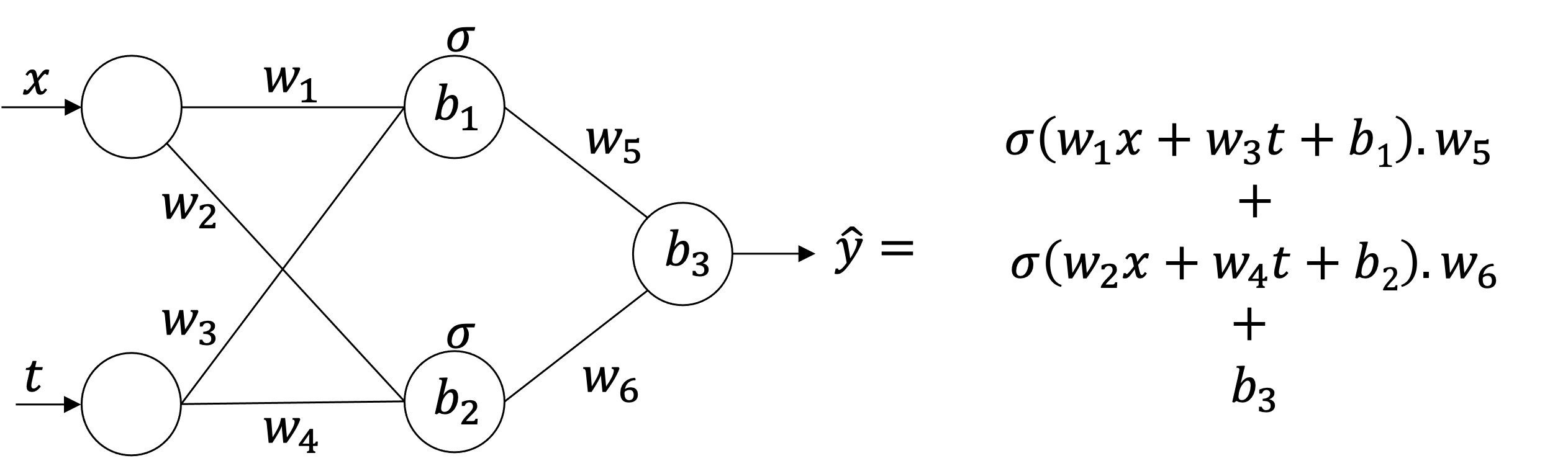}
 \caption{Architecture of the MLNN used in the toy example.}
 \label{Figure:MLNN-Arch}
\end{figure}

\textbf{Multi-layer neural network (MLNN) architecture:} For illustration, we use a basic MLNN with two inputs, one output, and one hidden layer, with 2 nodes as shown in Figure \ref{Figure:MLNN-Arch}. The hidden nodes and output nodes have an additional bias input. For simplicity, we have used sigmoid nodes in the hidden layers and output layer whereas the input layers are linear. For the readability of equations we use the following two notations: First, $f_{w_iw_jb_k} = \sigma(w_i.x + w_j.t + b_k) $ and second, $f_{w_i, b_k} = \sigma(w_i.x + b_k)$. For example, the notations simplify $\hat{y}$ to $w_5.f_{w_1w_3b_1} + w_6.f_{w_2w_4b_2} + b_3$.
\\The loss function in terms of network parameters is as follows, 
\begin{equation}
\begin{split}
 \mathcal{L} = \overbrace{\frac{1}{|T_f|} \sum_{(x, t)\in \mathcal{T}_f}\left[w_5.f_{w_1w_3b_1}.(1 - f_{w_1w_3b_1}).(3.w_1 + w_3) + w_6.f_{w_2w_4b_2}.(1 - f_{w_2w_4b_2}).(3.w_2 + w_4)\right]^2}^{\mathcal{L}_f}\\ + \underbrace{\frac{1}{|T_b|}\sum_{(x, t) \in \mathcal{T}_b}\left[w_5.f_{w_1b_1} + w_6.f_{w_2b_2} + b_3 - x.e^{-1*x^2}\right]^2}_{\mathcal{L}_b}.
\end{split}
\end{equation}

The weights of a given $x$ are updated using the back-propagation rule (Table \ref{Table:bp_wu}), where the loss function ($\mathcal{L}$) is used to update the weights as shown in Equations \ref{eq:dl_dw1}-\ref{eq:dl_dw9}. For readability let the equation in the first square bracket be denoted as $A$ and let the equation in the second square bracket be denoted as $B$.

In theory, when the weights are updated, $\hat{y}$ should approach the theoretical value of the function represented by the PDE for a given input $x$. Table \ref{table:pde_manual} shows the results for 5 iterations (i.e., loops) of $x=0.1$. It should be noted that the proposed architecture may not be the most appropriate for the given PDE, but is useful for a simple illustration of the PINN approach.
\setlength{\belowdisplayskip}{0pt} \setlength{\belowdisplayshortskip}{0pt}
\setlength{\abovedisplayskip}{0pt} \setlength{\abovedisplayshortskip}{0pt}
\begin{table}[htp]
\small
\centering
\caption{Back-propagation weight update rule for the weights.}
\label{Table:bp_wu}
\begin{tabular}{|p{1.75cm} | p{1.75cm} | p{1.75cm} | p{1.75cm}| p{1.75cm} | p{1.75cm}|}
 \hline
 $\mathbf{w_1}$ & $\mathbf{w_2}$ & $\mathbf{w_3}$ & $\mathbf{w_4}$ & $\mathbf{b_1}$ & $\mathbf{b_2}$ \\ 
 \hline
$w_1 + \eta.\frac{\partial \mathcal{L}}{\partial w_1}.x$ & $w_2 + \eta.\frac{\partial \mathcal{L}}{\partial w_2}.x$ & $w_3 + \eta.\frac{\partial \mathcal{L}}{\partial w_3}.x$ & $w_4 + \eta.\frac{\partial \mathcal{L}}{\partial w_4}.x$ & $b_1 + \eta.\frac{\partial \mathcal{L}}{\partial b_1}$ & $b_2 + \eta.\frac{\partial \mathcal{L}}{\partial b_2}$ \\ 
\hline 
 \multicolumn{2}{|c}{$\mathbf{w_5}$} & \multicolumn{2}{|c|}{$\mathbf{w_6}$} & \multicolumn{2}{|c|}{$\mathbf{b_3}$} \\ \hline
 \multicolumn{2}{|c}{$w_5 + \eta.\frac{\partial \mathcal{L}}{\partial w_5}.\sigma(w_1.x + w_3.t + b_1)$} & \multicolumn{2}{|c|}{$w_6 + \eta.\frac{\partial \mathcal{L}}{\partial w_6}.\sigma(w_2.x + w_4.t + b_2)$} & \multicolumn{2}{|c|}{$b_3 + \eta.\frac{\partial \mathcal{L}}{\partial b_3}$} \\ \hline
\end{tabular}
\label{table:pde_manual}
\end{table}

\begin{subequations}
\footnotesize
\begin{align}
\frac{\partial \mathcal{L}_f}{\partial w_1} &= \frac{2}{|T_f|}.A.\left[x.w_5.(3.w_1 + w_3).f_{w_1w_3b_1}.(1 - f_{w_1w_3b_1}).(1 - 2.f_{w_1w_3b_1}) + 3.w_5.f_{w_1w_3b_1}.(1 - f_{w_1w_3b_1}) \right],\\
\frac{\partial \mathcal{L}_b}{\partial w_1} &= \frac{2}{|T_b|}.B.\left[x.w_5.f_{w_1b_1}.(1 - f_{w_1b_1})\right].
\end{align}
\label{eq:dl_dw1}
\end{subequations}

\begin{subequations}
\footnotesize
\begin{align}
\frac{\partial \mathcal{L}_f}{\partial w_2} &= \frac{2}{|T_f|}.A.\left[ x.w_6.(3.w_2 + w_4).f_{w_2w_4b_2}.(1 - f_{w_2w_4b_2}).(1 - 2.f_{w_2w_4b_2}) + 3.w_6.f_{w_2w_4b_2}.(1 - f_{w_2w_4b_2})\right],\\
\frac{\partial \mathcal{L}_b}{\partial w_2} &= \frac{2}{|T_b|}.B.\left[x.w_6.f_{w_2b_2}.(1 - f_{w_2b_2})\right].
\end{align}
\end{subequations}

\begin{subequations}
\footnotesize
\begin{align}
\frac{\partial \mathcal{L}_f}{\partial w_3} &=  \frac{2}{|T_f|}.A.\left[ t.w_5.(3.w_1 + w_3).f_{w_1w_3b_1}.(1 - f_{w_1w_3b_1}).(1 - 2.f_{w_1w_3b_1}) + w_5.f_{w_1w_3b_1}.(1 - f_{w_1w_3b_1}) \right],\\
\frac{\partial \mathcal{L}_b}{\partial w_3} &= 0.
\end{align}
\end{subequations}

\begin{subequations}
\footnotesize
\begin{align}
\frac{\partial \mathcal{L}_f}{\partial w_4} &= \frac{2}{|T_f|}.A.\left[ t.w_6.(3.w_2 + w_4).f_{w_2w_4b_2}.(1 - f_{w_2w_4b_2}).(1 - 2.f_{w_2w_4b_2}) + w_5.f_{w_2w_4b_2}.(1 - f_{w_2w_4b_2}) \right],\\
\frac{\partial \mathcal{L}_b}{\partial w_4} &= 0.
\end{align}
\end{subequations}

\begin{subequations}
\footnotesize
\begin{align}
\frac{\partial \mathcal{L}_f}{\partial b_1} &= \frac{2}{|T_f|}.A.\left[ w_5.(3.w_1 + w_3).f_{w_1w_3b_1}.(1 - f_{w_1w_3b_1}).(1 - 2.f_{w_1w_3b_1}) \right],\\
\frac{\partial \mathcal{L}_b}{\partial b_1} &= \frac{2}{|T_b|}.B.\left[ w_5.f_{w_1b_1}.(1 - f_{w_1b_1}) \right].
\end{align}
\end{subequations}

\begin{subequations}
\footnotesize
\begin{align}
\frac{\partial \mathcal{L}_f}{\partial b_2} &= \frac{2}{|T_f|}.A.\left[ w_6.(3.w_2 + w_4).f_{w_2w_4b_2}.(1 - f_{w_2w_4b_2}).(1 - 2.f_{w_2w_4b_2}) \right],\\
\frac{\partial \mathcal{L}_b}{\partial b_2} &= \frac{2}{|T_b|}.B.\left[ w_6.f_{w_2b_2}.(1 - f_{w_2b_2})\right] .
\end{align}
\end{subequations}

\begin{subequations}
\footnotesize
\begin{align}
\frac{\partial \mathcal{L}_f}{\partial w_5} &= \frac{2}{|T_f|}.A.\left[ f_{w_1w_3b_1}.(1 - f_{w_1w_3b_1}).(3.w_1 + w_3) \right],\\
\frac{\partial \mathcal{L}_b}{\partial w_5} &=  \frac{2}{|T_b|}.B.\left[ f_{w_1b_1}\right].
\end{align}
\end{subequations}

\begin{subequations}
\footnotesize
\begin{align}
\frac{\partial \mathcal{L}_f}{\partial w_6} &= \frac{2}{|T_f|}.A.\left[ f_{w_2w_4b_2}.(1 - f_{w_2w_4b_2}).(3.w_2 + w_4) \right],\\
\frac{\partial \mathcal{L}_b}{\partial w_6} &= \frac{2}{|T_b|}.B.\left[ f_{w_2b_2}\right].
\end{align}
\end{subequations}

\begin{subequations}
\footnotesize
\begin{align}
\frac{\partial \mathcal{L}_f}{\partial b_3} &= 0,\\
\frac{\partial \mathcal{L}_b}{\partial b_3} &= \frac{2}{|T_b|}.B.
\end{align}
\label{eq:dl_dw9} 
\end{subequations}

\begin{table}[htp]
\footnotesize
\centering
\caption{Weight updates for the MLNN using manual calculation for $x=0.1, t=0.1$ with a learning rate of $0.05$. The solution of the PDE at the sample point is $-0.198$.}
\label{tab:bp_weights_manual}
\begin{tabular}{|p{0.6cm}|p{0.7cm}|p{0.7cm}|p{0.7cm}|p{0.7cm}|p{0.7cm}|p{0.7cm}|p{0.9cm}|p{0.9cm}|p{0.9cm}|p{0.7cm}|p{0.7cm}|}
\hline
\textbf{Loop} & $\mathbf{w_1}$ & $\mathbf{w_2}$ & $\mathbf{w_3}$ & $\mathbf{w_4}$ & $\mathbf{w_5}$ & $\mathbf{w_6}$ & $\mathbf{b_1}$ & $\mathbf{b_2}$ & $\mathbf{b_3}$ &  $\hat{y}$ & Loss\\
\hline
$0$ & $0.5$ & $0.5$ & $0.5$ & $0.5$ & $0.5$ & $0.5$ & $0$ & $0$ & $0$ & $0.525$ & $0.146$\\
$1$ & $0.495$ & $0.495$ & $0.503$ & $0.503$ & $0.464$ & $0.464$ & $-0.002$ & $-0.002$ & $-0.056$ & $0.665$ & $0.132$\\
$2$ & $0.492$ & $0.492$ & $0.505$ & $0.505$ & $0.434$ & $0.434$ & $-0.002$ & $-0.002$ & $-0.102$ & $0.571$ & $0.1008$\\
$3$ & $0.491$ & $0.491$ & $0.507$ & $0.507$ & $0.407$ & $0.407$ & $-0.003$ & $-0.003$ & $-0.139$ & $0.493$ & $0.0786$\\
$4$ & $0.491$ & $0.491$ & $0.509$ & $0.509$ & $0.384$ & $0.384$ & $-0.002$ & $-0.002$ & $-0.170$ & $0.427$ & $0.063$\\
$5$ & $0.493$ & $0.493$ & $0.511$ & $0.511$ & $0.364$ & $0.364$ & $-0.002$ & $-0.002$ & $-0.194$ & $0.372$ & $0.052$\\
\hline
\end{tabular}
\end{table}

\begin{table}[htp]
\small
\centering
\caption{Weight updates for the MLNN using DeepXDE module for $x=0.1, t=0.1$. The solution of the PDE at the sample point is $-0.198$.}
\label{tab:bp_weights_deepxde}
\begin{tabular}{|p{0.6cm}|p{0.7cm}|p{0.7cm}|p{0.7cm}|p{0.7cm}|p{0.7cm}|p{0.7cm}|p{0.9cm}|p{0.9cm}|p{0.9cm}|p{0.7cm}|p{0.7cm}|}
\hline
\textbf{Loop} & $\mathbf{w_1}$ & $\mathbf{w_2}$ & $\mathbf{w_3}$ & $\mathbf{w_4}$ & $\mathbf{w_5}$ & $\mathbf{w_6}$ & $\mathbf{b_1}$ & $\mathbf{b_2}$ & $\mathbf{b_3}$ &  $\hat{y}$ & Loss\\
\hline
$0$ & $0.5$ & $0.5$ & $0.5$ & $0.5$ & $0.5$ & $0.5$ & $0$ & $0$ & $0$ & $0.5249$ & $1.00$\\
$1$ & $0.499$ & $0.499$ & $0.499$ & $0.499$ & $0.499$ & $0.499$ & $-0.001$ & $-0.001$ & $-0.001$ & $0.5226$ & $0.996$\\
$2$ & $0.4970$ & $0.4970$ & $0.4970$ & $0.4970$ & $0.4970$ & $0.4970$ & $-0.0029$ & $-0.0029$ & $-0.0029$ & $0.5179$ & $0.981$\\
$3$ & $0.4940$ & $0.4940$ & $0.4940$ & $0.4940$ & $0.4940$ & $0.4940$ & $-0.0059$ & $-0.0059$ & $-0.0059$ & $0.5109$ & $0.960$\\
$4$ & $0.4900$ & $0.4900$ & $0.4900$ & $0.4900$ & $0.4900$ & $0.4900$ & $-0.0099$ & $-0.0099$ & $-0.0099$ & $0.5015$ & $0.932$\\
$5$ & $0.4850$ & $0.4850$ & $0.4850$ & $0.4850$ & $0.4850$ & $0.4850$ & $-0.0149$ & $-0.0149$ & $-0.0149$ & $0.4899$ & $0.897$\\
\hline
\end{tabular}
\end{table}

In this example, the equation is a transport equation whose general form is given by $\frac{\partial u}{\partial t} + v.\frac{\partial u}{\partial x} = 0$, where $v$ is a constant. The exact solution to the equation in the example is $u(x,t) = (x - 3t).e^{-(x-3t)^2}$. (Proving the exact solution is beyond the scope of this paper). 



\subsection{Implementation of the PINN DE Example}
\label{dnn_pde_example}
The main body of code presented in Raissi et al. \cite{Raissi2017} has been widely used in a number of studies \cite{Tzen2019, Kharazmi2021, Lu2021}. To adapt the original code for general engineering challenges, several wrapper libraries have been established. In this section, the python modules developed in Lu et al. \cite{Lu2019} are examined. The network is trained with different hyperparameters  to demonstrate its effects on the efficiency of the model's prediction. This section also includes a tutorial for the implementation of a PINN PDE solver to solve the PDE discussed in Section \ref{App:PINN_EG}.

\subsubsection*{Forward Problem - Hyperparameter modulation} \label{App:p0-forward}
Section \ref{App:PINN_EG} introduced an example of a forward problem that can be effectively used to demonstrate the implementation of a DNN to solve a PDE. For convenience, the problem is formulated again as follows:
\\\textbf{Main function} ($F$) = $(x - 3t).e^{-(x-3t)^2}$
\\\textbf{Partial Differential Equation} ($PDE$) : $\frac{\partial \hat{y}}{\partial t} + 3\frac{\partial \hat{y}}{\partial x}$
\\\textbf{Initial Condition} $u(x,0) = x.e^{-x^2}$ \par
\begin{figure}[htp]
 \centering
 \includegraphics[width=0.6\linewidth]{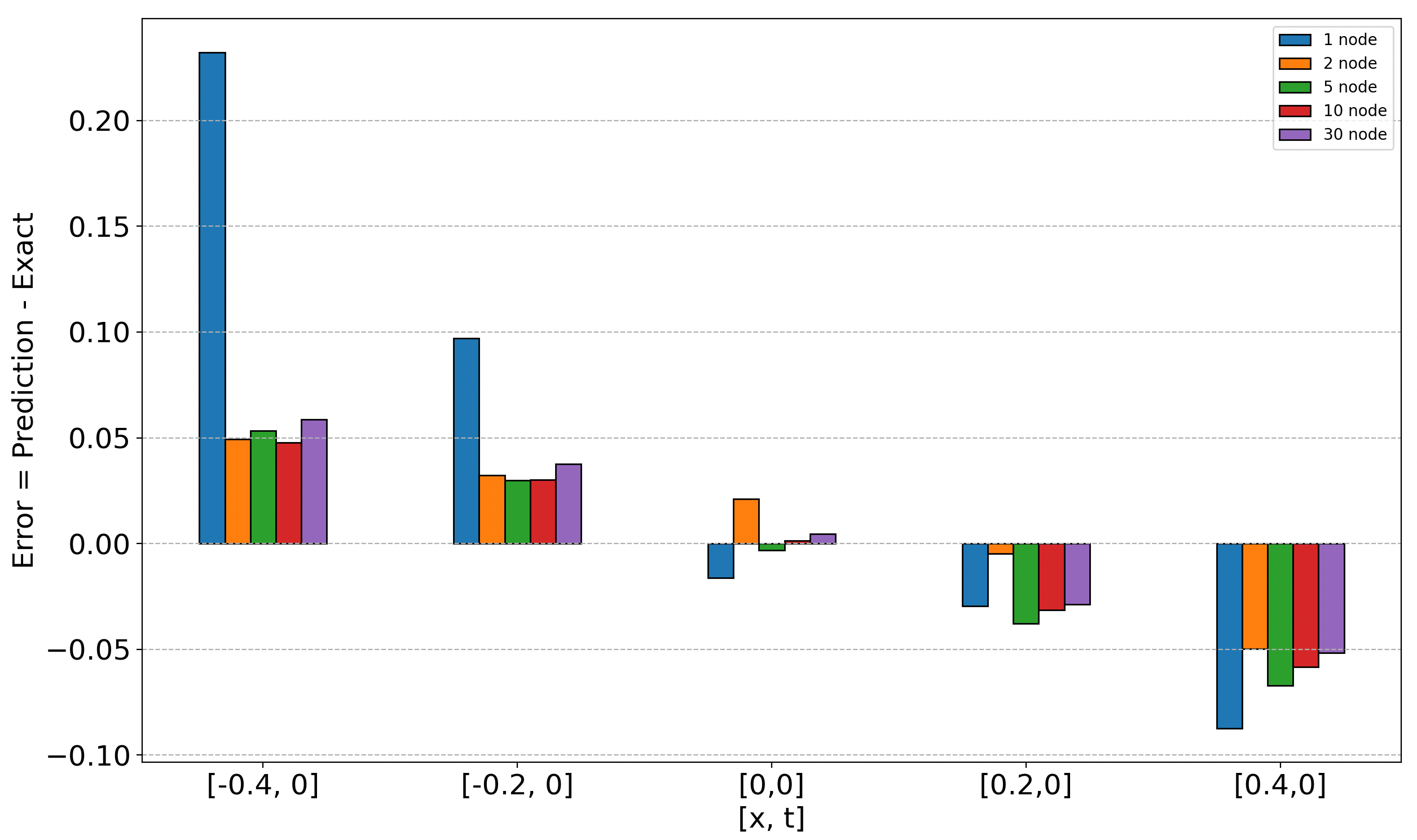}
 \caption{Prediction error with varying number of nodes in the hidden layer of the network}
 \label{Figure:error_varying_nodes}
\end{figure}

Prior to beginning the generalised tutorial for the implementation of a PINN PDE solver, a series of initial tests was carried out with the aim of demonstrating the workings of the python module under discussion. The first involved starting off with a pre-determined network architecture and observing how the network performed on varying either the number of nodes or the depth of the network, keeping the other hyperparameters fixed. Initially, a network with a single hidden layer was chosen and the number of nodes in the layer was varied from 1 to 30 in non-uniform intervals. Figure \ref{Figure:error_varying_nodes} displays the prediction errors at 5 different data points. It can be observed that on an average, networks with $10$ or $30$ nodes perform better than networks with lesser number of nodes, with an exception of the network with $2$ nodes which gives a lower error at certain points and a higher error at certain points. 

\begin{figure*}[h!]
     \centering
     \begin{subfigure}[b]{0.45\linewidth}
         \includegraphics[width=\linewidth]{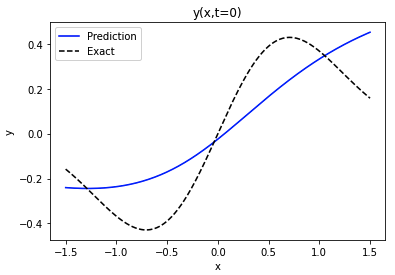}
         \caption{}
     \end{subfigure}
     \begin{subfigure}[b]{0.45\linewidth}
         \includegraphics[width=\linewidth]{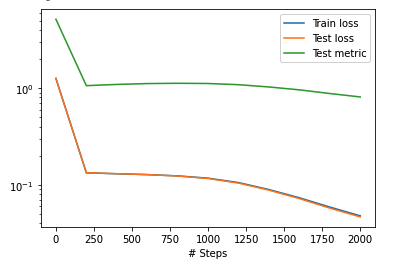}
         \caption{}
     \end{subfigure}
     \caption{Model prediction and convergence curve for learning rate of 0.001}
      \label{fig:1_layer_10000epoch_0.001lr}
    \end{figure*}

\begin{figure*}[h!]
     \centering
     \begin{subfigure}[b]{0.45\linewidth}
         \includegraphics[width=\linewidth]{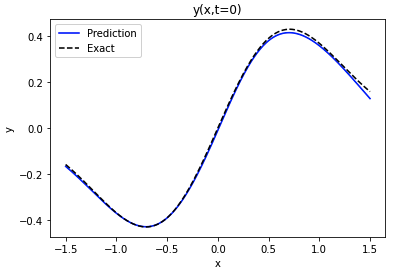}
     \end{subfigure}
     \begin{subfigure}[b]{0.45\linewidth}
         \includegraphics[width=\linewidth]{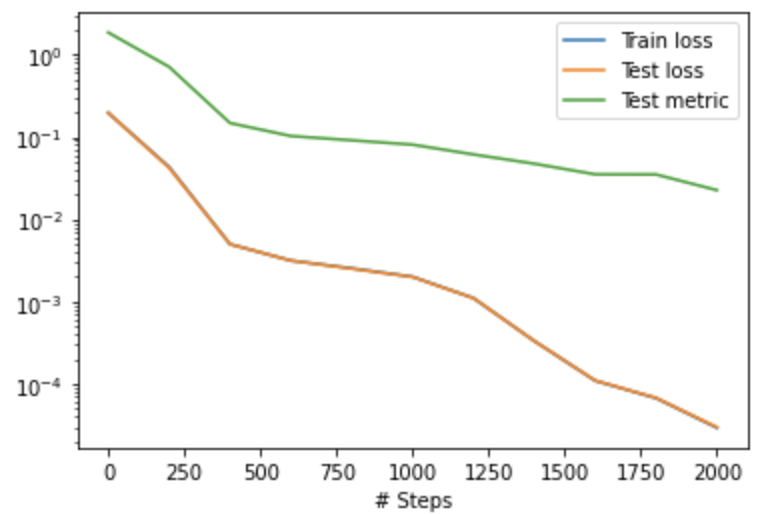}
     \end{subfigure}
     \caption{Model prediction and convergence curve for learning rate of 0.1}
      \label{fig:1_layer_10000epoch_0.1lr}
    \end{figure*}

Two training simulations were then carried out to help visualize the trends set by varying the learning rate. A single hidden layer model with 32 nodes was trained for 2000 epochs first with a learning rate of 0.001  and then with a learning rate of 0.1. Figure \ref{fig:1_layer_10000epoch_0.001lr} and Figure \ref{fig:1_layer_10000epoch_0.1lr} illustrate the curve fitting performance and the convergence of the model respectively. The model works as expected: with a slow learning rate (0.001) we see a slower convergence and a poor fit relative to the performance of the model at a higher learning rate (0.1). 

\begin{figure*}[b!]
     \centering
     \begin{subfigure}[b]{0.45\linewidth}
         \includegraphics[width=\linewidth]{tutorialfigs/forward_2000epoch_0.001lr_40samplesize.png}
         \caption{Model prediction at 2000 epochs}
     \end{subfigure}
     \begin{subfigure}[b]{0.45\linewidth}
         \includegraphics[width=\linewidth]{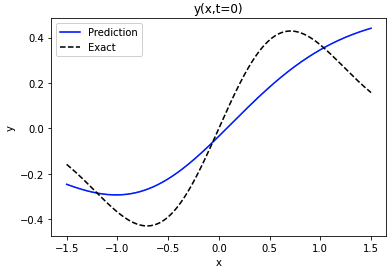}
         \caption{Model prediction at 3000 epochs}
     \end{subfigure}
      \begin{subfigure}[b]{0.45\linewidth}
         \includegraphics[width=\linewidth]{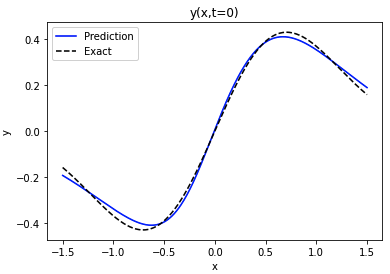}
         \caption{Model prediction at 4000 epochs}
     \end{subfigure}
      \begin{subfigure}[b]{0.45\linewidth}
         \includegraphics[width=\linewidth]{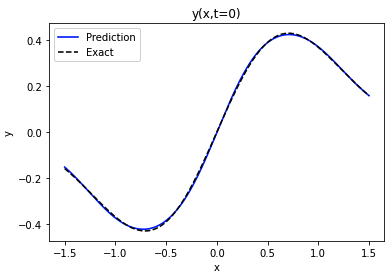}
         \caption{Model prediction at 5000 epochs}
     \end{subfigure}
     \caption{Curve fitting evolution for learning rate of 0.001}
     \label{fig:2_layer_0.001lr}
    \end{figure*}

Next, simulations to help visualize the trends in the curve-fitting capability at various epochs were carried out. A 2-layer model with 64 nodes per layer was trained for 2000, 3000, 4000, and 5000 epochs with a learning rate of 0.001 in order to illustrate how the curve-fitting performance improves through epochs. Figure \ref{fig:2_layer_0.001lr} shows the evolution of the curve fitting performance of the model. From these figures, a distinct reduction in the prediction error (equivalent to an increased curve-fitting capability of the model) can be observed.

\subsubsection*{Implementation tutorial - Forward Problem} \label{App:forward tutorial} 
In this section, we demonstrate how to implement the DeepXDE package to solve the forward problem introduced in Section \ref{App:PINN_EG}.
\begin{itemize}
    \item First, we import the necessary python packages (DeepXDE). 
    \begin{figure}[H]
        \centering
        \includegraphics[width=0.8\linewidth]{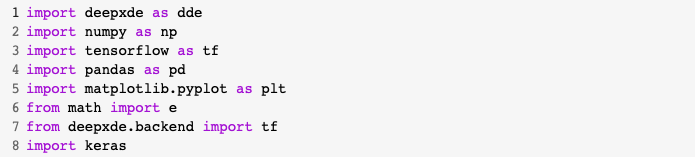}
    \end{figure}
    \item We then define the PDE that we aim to solve. We also define the function that represents the exact solution of the PDE. The exact solution is only defined for running the simulations as shown earlier. This step would not be required to solve a forward problem.
    \begin{figure}[H]
        \centering
        \includegraphics[width=0.8\linewidth]{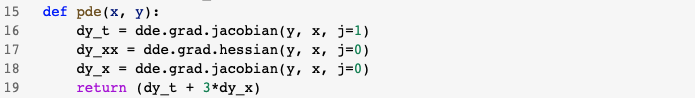}
    \end{figure}
    \item We then define the computational geometry and the time domain that will be used for the training process. The built-in classes \texttt{Interval}, \texttt{TimeDomain} and \texttt{GeometryXTime} can be used as shown below.
    \begin{figure}[H]
        \centering
        \includegraphics[width=0.8\linewidth]{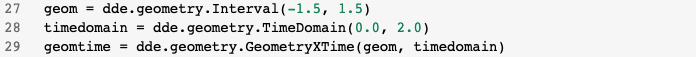}
    \end{figure}
    \item Next, we consider the boundary/initial conditions. The boundary condition is set to 0 for this example. The initial condition is defined using the class \texttt{IC} as shown below.
    \begin{figure}[H]
        \centering
        \includegraphics[width=0.8\linewidth]{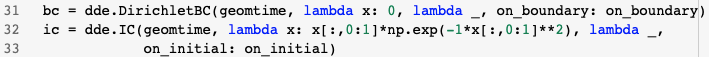}
    \end{figure}
    \item Now we define the \texttt{TimePDE} forward problem as shown below. The number 8190 is the number of residual training  points sampled inside the domain, and the number 4094 is the number of training points sampled on the boundary. We also include 4094 initial residual points for the initial conditions. The numbers are arbitrarily chosen as deemed fit.
    \begin{figure}[H]
        \centering
        \includegraphics[width=0.8\linewidth]{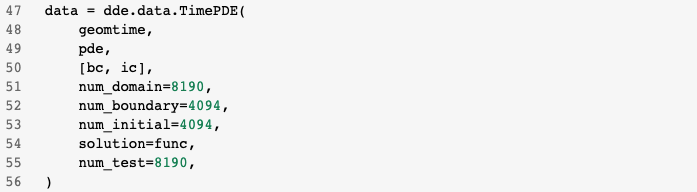}
    \end{figure}
    \item Next, we define the network architecture. Here, we use a fully connected neural network of depth 3 (i.e., 2 hidden layers) and width 64. We have chosen a sigmoid function activated network.
    \begin{figure}[H]
        \centering
        \includegraphics[width=0.8\linewidth]{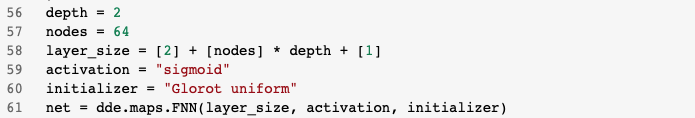}
    \end{figure}
    \item We then build a model, choose an optimizer, fix the learning rate, specify the loss metric that we use to assess the performance, define the number of training epochs and finally train the model using the syntax as shown below.
    \begin{figure}[H]
        \centering
        \includegraphics[width=0.8\linewidth]{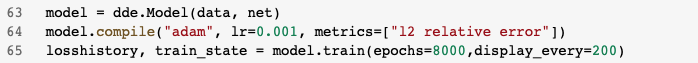}
    \end{figure}
    \item Post training, we define the input over which we wish to predict using the trained model.
    \begin{figure}[H]
        \centering
        \includegraphics[width=0.8\linewidth]{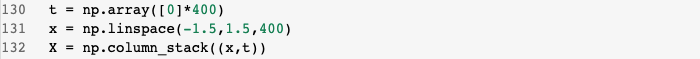}
    \end{figure}
    \item Finally, we predict the testing outputs for the inputs defined in step 8 and print the required output.
    \begin{figure}[H]
        \centering
        \includegraphics[width=0.8\linewidth]{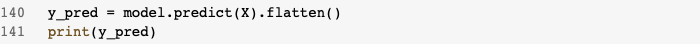}
    \end{figure}

\end{itemize}

\subsubsection*{Inverse Problem - Hyperparameter modulation}
To demonstrate the capabilities of a DNN-DE solver for identifying the coefficients of a partial differential equation, experiments and a tutorial for the inverse problem, similar to the one presented in Section \ref{App:forward tutorial}, is presented in this and the following section respectively. To maintain uniformity and simplicity, the same transport equation is used to demonstrate the implementation of the DeepXDE package to solve inverse problems. We restructured the problem definition to that of an inverse problem by assuming the coefficient of the transport equation to be unknown (here, C). Hence, the problem is formulated as:
\\\textbf{Main function} ($F$) = $(x - 3t).e^{-(x-3t)^2}$
\\\textbf{Partial Differential Equation} ($PDE$) : $\frac{\partial \hat{y}}{\partial t} + C\frac{\partial \hat{y}}{\partial x}$
\\\textbf{Initial Condition} $u(x,0) = x.e^{-x^2}$
\par
\begin{figure*}[h!]
     \centering
     \begin{subfigure}[b]{0.45\linewidth}
         \includegraphics[width=\linewidth]{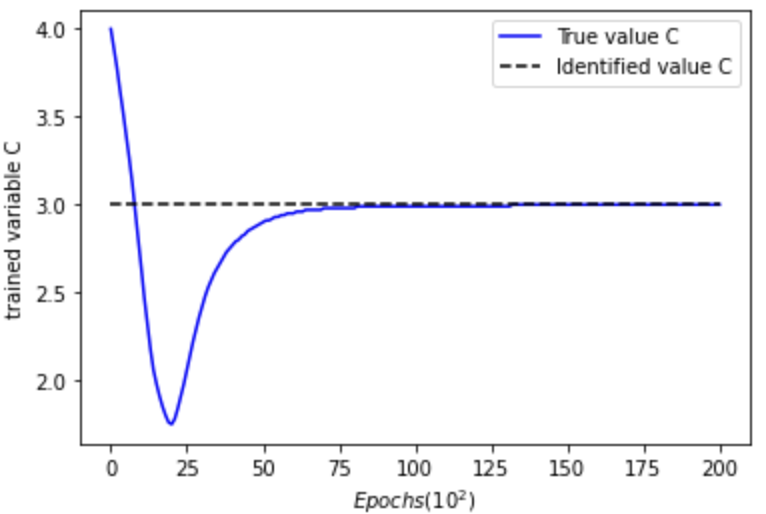}
     \end{subfigure}
     \begin{subfigure}[b]{0.45\linewidth}
         \includegraphics[width=\linewidth]{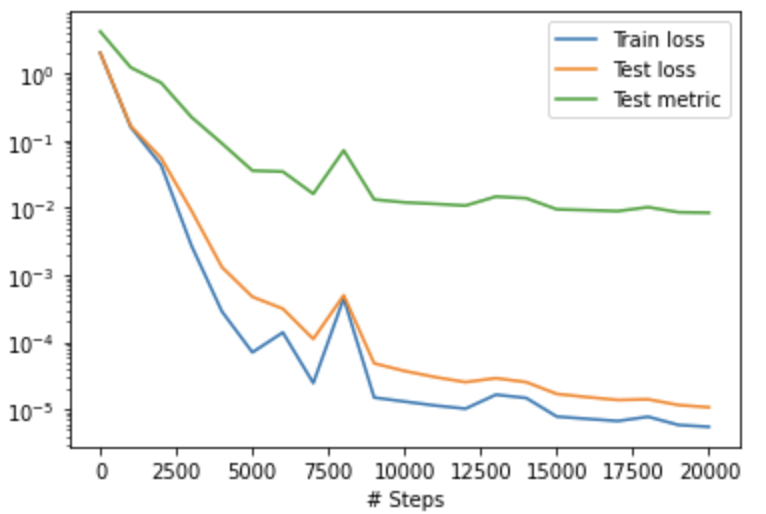}
     \end{subfigure}
     \caption{Variable prediction and Loss convergence at a learning rate of 0.001.}
      \label{fig:inverse_0.001lr_64_nodes}
    \end{figure*}
\begin{figure*}[h!]
     \centering
     \begin{subfigure}[b]{0.45\linewidth}
         \includegraphics[width=\linewidth]{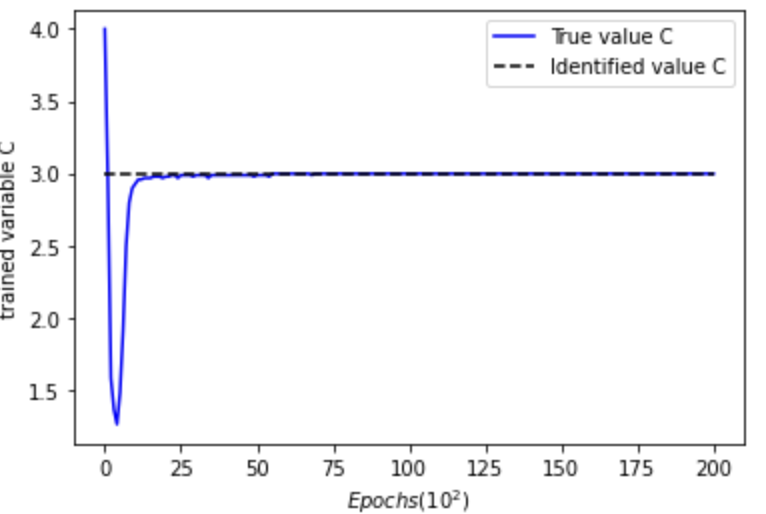}
     \end{subfigure}
     \begin{subfigure}[b]{0.45\linewidth}
         \includegraphics[width=\linewidth]{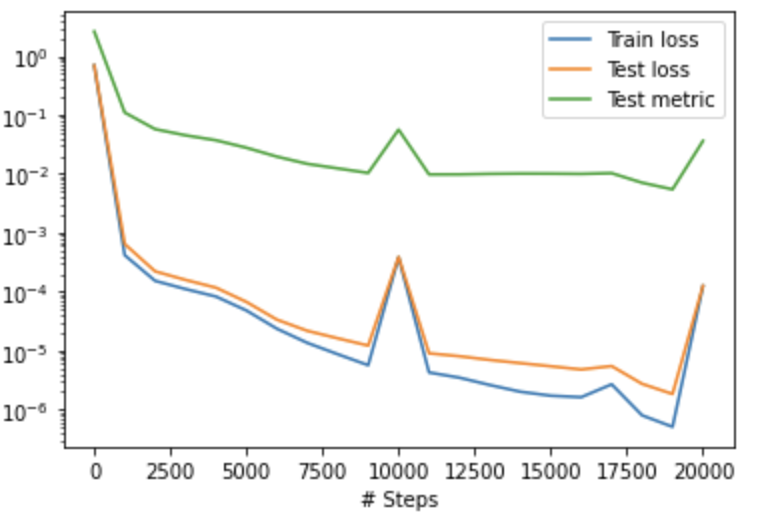}
     \end{subfigure}
     \caption{Variable prediction and Loss convergence at a learning rate of 0.01.}
      \label{fig:inverse_0.01lr_64_nodes}
    \end{figure*}

A series of test, similar to those that were run for the forward problem in Section \ref{App:p0-forward}, were carried out for the inverse problem. The performance of the model in solving the inverse problem was assessed by visualizing and quantifying how well the trained variable $C$ converges to its true value. A model with two hidden layers consisting of 64 nodes each was trained at a learning rate of 0.001 and 0.01 for 20000 epochs with the unknown variable $C$ being updated every 100 epochs. The optimizer used for this training process was the Adam optimizer. It can be observed from Figure \ref{fig:inverse_0.001lr_64_nodes} and Figure \ref{fig:inverse_0.01lr_64_nodes} that with increasing learning rate the convergence is significantly faster.


\subsubsection*{Implementation Tutorial - Inverse Problem}
In this section, we demonstrate how to implement the
DeepXDE package to solve the formulated inverse problem. 
\begin{itemize}
    \item The first four steps are identical to those of the forward problem: import the required modules, define the PDE, define the computational geometry and the time domain, and set the boundary and initial conditions.  It is to be noted that within the function that defines the PDE, the formulated equation is returned with the unknown variable $C$.
    \begin{figure}[H]
        \centering
        \includegraphics[width=0.8\linewidth]{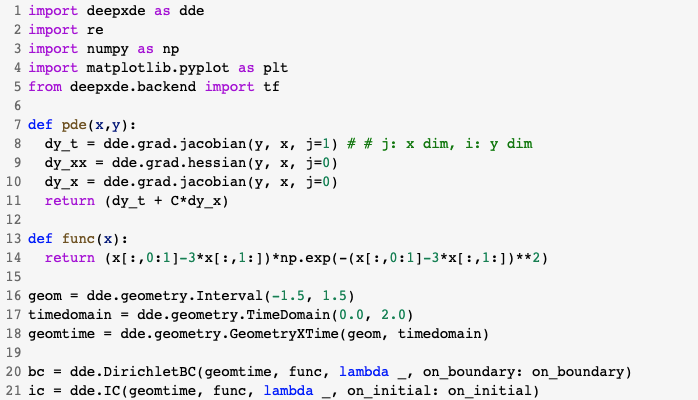}
    \end{figure}
    \item The unknown trainable variable is then initialized. For the purpose of performance analysis of the model, the true value of the unknown variable is also defined.
    \begin{figure}[H]
        \centering
        \includegraphics[width=0.8\linewidth]{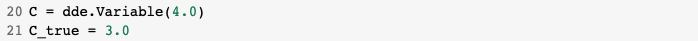}
    \end{figure}
    \item Additional information to help determine the value of the unknown variable C is then passed onto the model, where  $observe\_x$ is a 2-D array of equally spaced input points $(x,t)$; $observe\_y$ is a variable of the \texttt{PointSetBC} class that defines the Dirichlet boundary condition for a set of points; and  $func(observe\_x)$ represents the \textit{target} data points.
    \begin{figure}[H]
        \centering
        \includegraphics[width=0.8\linewidth]{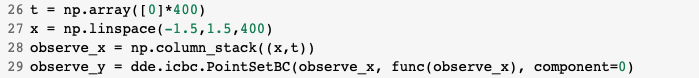}
    \end{figure}
    \item The TimePDE problem is then defined. It is necessary to additionally pass the extra boundary condition variables $observe\_y$ and $observe\_x$.
    \begin{figure}[H]
        \centering
        \includegraphics[width=0.8\linewidth]{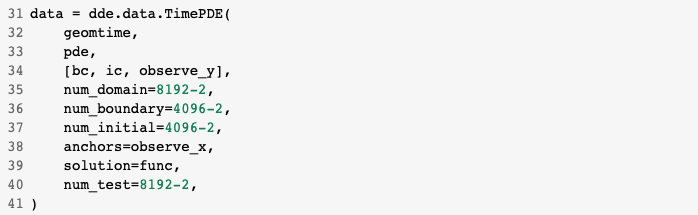}
    \end{figure}
    \item The network architecture, the activations and the initializers are then defined in the same way as for the forward problem.
    \begin{figure}[H]
        \centering
        \includegraphics[width=0.8\linewidth]{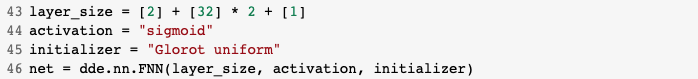}
    \end{figure}
    \item Next, the model is built, the optimizer is chosen, the learning rate is fixed, the loss metric is specified,  and the unknown variable is passed onto the model through the variable $external\_trainable\_variables$.
    \begin{figure}[H]
        \centering
        \includegraphics[width=0.8\linewidth]{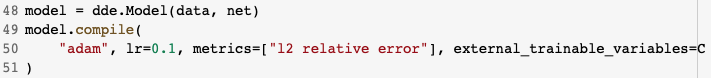}
    \end{figure}
    \item The values trainable variable $C$ is then set up to be stored in a file in real time at a specified frequency. The model is then trained for a specified number of epochs.
    \begin{figure}[H]
        \centering
        \includegraphics[width=0.8\linewidth]{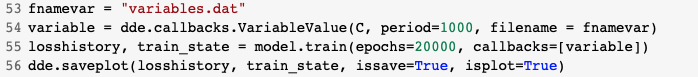}
    \end{figure}
    
\end{itemize}

\section{Discussion}
\label{sec:discussion}
In Section \ref{sec:discussion}, key lessons learned from the survey and computational example are shared. Section \ref{sec:discuss implem} summarizes experience setting up PINN's hyper-parameters in Section \ref{sec:tutorial}. In Section \ref{App:AD}, a high level overview of automatic differentiation is provided as a principle of that allows for DNNs to compute derivatives. Section \ref{sec:numerical_methods} provides a comparison between finite method and DNN-DE methods. Section \ref{sec:generalization error} highlights future work of formalizing error of using DNN-DE for either model solution or model discovery.

\subsection{Implementation of DeepXDE}
\label{sec:discuss implem}
The DeepXDE python package was chosen to demonstrate the capabilities of feed-forward deep neural networks to solve partial differential equations. The package incorporates a back-end code that is much shorter and more comprehensive than traditional numerical methods. It is made to be sufficiently sophisticated for solving PDEs as it has a fairly straightforward syntax to define the problem as a whole, including computational domain
(geometry and time), PDE equations, boundary/initial conditions, constraints, training data, neural network architecture, and training of the hyper-parameters \cite{Lu2021}.  

To illustrate the use of the package in solving a forward problem, we chose a transport equation. Executing this example involved manual tuning of numerous variables and hyper-parameters, the first of which was choosing an appropriate computational geometry and time domain within which the training of the model was to be performed. For simplicity, the assessment of the algorithm’s performance was planned to be done at $t=0$. Hence, an appropriate time domain of $[0.0, 2.0]$ was chosen. At $t=0$, the exact solution $(x-3t)e^{-(x-3t)^2}$ converges to 0 in the approximate domain $(-\infty, -2]\cup[2,\infty)$. Hence, $[-1.5, 1.5]$ was chosen as the computational geometry for the sampled training space. 

Choosing the sample size of the training dataset for the forward problem was fairly arbitrary: we chose to train the DNNs with sample sizes of lengths close to powers of $2$ which was considered a reasonable starting point. The same sample size was set for the number of data points sampled at the boundaries and for the initial conditions. However, the network displayed similar performance when trained with a sample size of much lesser magnitude ($40$ samples within the domain, $20$ samples on the boundaries and $10$ samples for the initial conditions) with much less training time $-2.23$ seconds for $2000$ epochs at a learning rate of $0.001$ with a smaller sample size of $40$ against $70.41$ seconds with the same hyper-parameters and a larger sample size of $8190$. This indicates that for the domain length that we address in this problem, the network does well with sample sizes as small as $40$. 

The DeepXDE package provides numerous optimizer options to choose from, including a stochastic gradient descent (SGD), a momentum optimizer, and an Adam optimizer. For the purpose of creating Table \ref{tab:bp_weights_deepxde}, the optimizer that works on a similar set of weight update rules as shown in Table \ref{Table:bp_wu} would be the best choice, and hence, SGD, which is a special case of vanilla GD (gradient descent) where the batch size is $1$, was chosen. However, for the purpose of training a neural network to fit the exact solution of a PDE, in the forward problem, the optimizer that accelerates the training process the most would be the best choice. The Adam optimizer was chosen for the training process as it combines the accelerating capabilities of the momentum optimizer and the RMSProp optimizer. It stores the individual learning rate of RMSProp and the weighted average of the gradients from the momentum optimizer. The sigmoid function was chosen as the activation function to keep things simple. The model did not show any significant change in performance when trained on other activation functions. The process of parameter initialization for the training process was executed using the Glorot uniform initializer as it ensures that the mean of the activations is zero and the variance of the activations is equal across layers. This allows the back-propagation gradient signal to travel through the layers without exploding (due to large initializations) or vanishing (due to small initializations). However, to create Table \ref{tab:bp_weights_deepxde}, a constant (equal to 0.5) initialization was used in order to maintain the similarity between the model equations and the manually calculated values. 

The same transport equation was chosen to illustrate the use of the package in solving an inverse problem. One of the known coefficients of the equation was assumed to be unknown. Similar to the forward problem example, the time domain was set as $[0.0,2.0]$ and the computational geometry was set as $[-1.5,1.5]$. From an implementation point of view, appropriate time domain and computational geometry can be chosen based on the available observations of the independent variables that the model is going to be trained on. The training sample size, the number of data points sampled at the boundaries and the number of data points sampled for the initial conditions were all chosen in a similar manner as with the forward problem. The model was trained with both the Adam optimizer and the SGD optimizer. During several training runs with the SGD optimizer, the algorithm ran into converging errors where the value of the trained variable would not converge to the true value. By contrast, the Adam optimizer converged fairly well with the trained variable. Parameter initialization for the training process was executed using the Glorot uniform initializer for the same reason as mentioned earlier. 

\subsection{Automatic Differentiation}
\label{App:AD}
Principle of using DNNs for solving or discovering DEs is closely related to the property of Automatic differentiation (AD) of DNNs. AD is based on the fact that all numerical computations are compositions of a finite set of elementary operations for which derivatives are known. Combining the derivatives of the constituent operations using the chain rule gives the overall composition derivative. In general, there are forward and reverse accumulating modes \cite{Lu2021} of automatic differentiation.

We can see that AD requires only one forward pass and one backward pass to compute all the partial derivatives, no matter what the input dimension is. In contrast, using finite differences to compute each partial derivative $\frac{\partial y }{\partial x_i}$ requires two function valuations $y(x_1,...,x_i,...,x_{d_{in}})$ and $y(x_1,...,x+i + \Delta x+i,...,x_{d_{in}})$ for some small number $\Delta x_i$, and thus in total $d_{in} + 1$ forward passes are required to evaluate all the partial derivatives. Hence, AD is much more efficient than finite difference when the input dimension is high (see \cite{baydin2018automatic, margossian2019review} for more details of the comparison between AD and other methods). To compute $n^th$-order derivatives, AD can be applied recursively $n$ times. However, this nested approach may lead to inefficiency and numerical instability, and hence other methods, e.g., Taylor-mode AD, have been developed for this purpose \cite{betancourt2018geometric, bettencourt2019taylor}. Finally we note that with AD we differentiate the neural network and therefore we can deal with noisy data \cite{pang2019fpinns}.



\subsection{Comparison to Numerical Methods}
\label{sec:numerical_methods}
While neural network-based DE solvers have shown a strong capacity to provide fast approximate solutions to DEs, numerical methods are often employed to retrieve the ground truth solutions when evaluating DNN-DE performance. The primary advantages of DNN-DEs are (i) their relative speed, especially at solving problems involving high-dimensional parameter spaces, (ii) their wide applicability to different forward and inverse problems, (iii) their ability to solve complex problems via supervised or unsupervised learning, and (iv) their generalization to cases unseen during training, which would ordinarily require separate numerical simulations to evaluate. These advantages generally come at the cost of some relative error between the DNN-DE prediction and the ground truth solution from numerical methods.

Gao et al. compared their DNN-DE architecture to PINNs, and used finite volume-based numerical simulations to attain ground truth solutions, with their proposed DNN-DE method showing lower relative error and faster convergence than a PINN with the same training resources \cite{Gao2021}. Hesthaven et al. compared their own DNN-DE method with Galerkin methods, using FEMs to compute relative errors \cite{Hesthaven2018}. The Deep Galerkin Method proposed by Sirignano et al. showed very low absolute and relative errors compared to the finite difference method (FDM) solutions for multiple high-dimensional PDEs \cite{Sirignano2018}. Li et al. reported low relative errors compared to FEM and FDM-derived solutions when compared to their Fourier neural operator and multiple other DNN-DE methods \cite{li2020fourier}. Mishra et al. showed well-trained PINNs to be highly generalizable, and reported low relative error on multiple complex systems with known exact solutions, pointing to methodological differences between DNN-DEs and numerical methods rather than comparing them empirically \cite{Mishra2021}.

In each of these works, despite solving complex and high-dimensional PDEs, it was rare to see a relative error above 10\% even when the network was trained in a completely unsupervised manner, applied to a complex high-dimension problem, or tested with out-of-distribution inputs. Relative errors generally improved with additional training resources, typically falling below 2\%, and often below 1\%, across large parameter spaces for a number of problems. The time to solve a PDE using DNN-DE methods was typically reported to be at least one order of magnitude faster than traditional numerical solvers, but showed even more favorable speed-ups for inverse problems. Furthermore, traditional methods must run for each instance of a problem while DNN-DE methods, once training is complete, offer fast performance for a wide range of input conditions (in the forward case) and observations (in the inverse case) without sacrificing much accuracy.


\subsection{Formalizing Performance and Error of DNN-DE}
\label{sec:generalization error}
\begin{figure}[htp]
 \includegraphics[width=0.75\linewidth]{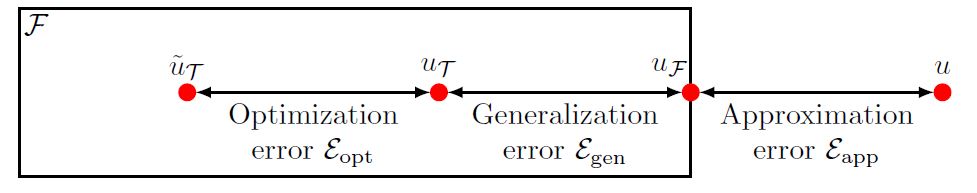}
 \caption{Illustration of errors of a parametric DNN-DE methods. $\mu$ denotes the PDE solution, $\mu_\mathcal{F}$ the best function closest to $\mu$ in the function space $\mathcal{F}$, $\mu_\mathcal{T}$ the neural network whose loss is at a global minimum, and $\tilde{\mu}_\mathcal{T}$ the function obtained by training a neural network. Unlike solution-driven DNN-DE, generalization error is added. \cite{Lu2021}}
 \label{fig:generalization_error}
 \centering
\end{figure}

Deducing a differential equations system based on observations in a hindsight risks over-fitting the model or missing the observation space completely. Significant interest from the mathematics and computer science community has started in the area to quantify this modeling error. Figure \ref{fig:generalization_error} illustrates a breakdown of the total error between true observation $\mu$ and an estimation based on the approximated model, $\tilde{\mu}_\mathcal{T}$. 

Optimization error, $\mathcal{E}_{opt}$ stems from the loss function complexity and
the optimization setup, such as learning rate and number of iterations. Approximation error, $\mathcal{E}_{app}$ indicates error between what the model represents and the real world phenomenon it tries to capture. Generalization error, $\mathcal{E}_{gen}$, deserves a special attention in this paper. This error represents the error due to the discrepancy between the inversely learned model and the DE it is training to approximate. Because one can formalized a DE and quantify the difference between the generalized and true systems, $\mathcal{E}_{gen}$ is a unique measure of how well a DNN performs. 

Mishra et. al in \cite{Mishra2021} gives mathematical overview of quantifying generalization error of PINN, as designed in \cite{Raissi2019}. The takeaway of \cite{Mishra2021} based on its empirical case studies of several complex systems (Poisson, heat transfer, flow dynamics) that on the basis of accuracy and speed, DNN-DE's performance in solving a model is superior to that of FEM and that DNN-DE can inverse high dimensional problems with a significant accuracy is also noteworthy.

Future work remains to further improve the quantification of DNN-DE's modeling error, with a holistic examination between the three components of errors identified in \ref{fig:generalization_error}. For instance, non-linear DE models present challenges in quantifying both $\mathcal{E}_{opt}$ and $\mathcal{E}_{app}$, making the evaluation of $\mathcal{E}_{gen}$ less certain nor substantive.

\section{Conclusions and Future Works}
\label{sec:conclusions}
This paper provided an overview of mathematical models and the methods with which to solve them. Literature delving in DNN-DE as an learning-based alternative method to the conventional numerical method was surveyed and classified based on the proposed taxonomy. A theoretical exploration of PINN, a popular architecture of DNN-DE, was provided, followed by a tutorial to implement a forward solver and a inverse modeler using DeepXDE, a Python package of PINN. Empirical results of the tutorial were presented as well as the experience of tuning the DeepXDE package's hyper parameters. 
Measuring DNN-DE's performance and error to ensure the reliability and transparency of the method is a major task beckoning future work. For solving a DE model, literature and preliminary computing example presented in this paper indicated there is a dominance zone where traditional numerical methods, such as finite element method, will give more accurate result. Solidifying the properties of a system that make DNN-DE  preferable, as well as the choice of a specific DNN-DE, is an important innovation that requires further research. Discovering a model with a formalized error estimation, which indicates how close the learned model is versus the desired system, is a recently growing field that shows potential for both the reliable applications of DNN-DE methods and the robust design of the methods.

\section*{Acknowledgement}
This material is based upon work supported by the National Science Foundation under Grant No. 1737633. We would like to thank Dr. Sairaj Dhople (University of Minnesota - Twin Cities) for providing useful feedback for the earlier draft of this manuscript. We would like to thank the spatial computing research group and Kim Koffolt for their helpful comments and refinements.

\bibliographystyle{ACM-Reference-Format}
\bibliography{sample-base.bib}

\end{document}